\documentclass[journal]{IEEEtran}

\IEEEoverridecommandlockouts

\usepackage[numbers]{natbib}
\usepackage{amsmath,amssymb,amsfonts}
\usepackage{algorithmic}
\usepackage{graphicx}
\usepackage{textcomp}
\usepackage{xcolor}
\usepackage{times}
\usepackage{soul}
\usepackage{url}
\usepackage[hidelinks]{hyperref}
\usepackage[utf8]{inputenc}
\usepackage[small]{caption}
\usepackage{multirow}
\usepackage{booktabs}
\usepackage{float}
\usepackage{subfigure}
\usepackage{graphicx}
\usepackage{bm}
\usepackage{enumerate}
\usepackage{color}
\usepackage{array}
\usepackage{amsthm}

\usepackage{algorithm, algorithmic}

\begin{document}

	\title{FDDH: Fast Discriminative  Discrete  Hashing for Large-Scale Cross-Modal Retrieval}
	
\author{Xin Liu,~\IEEEmembership{Senior Member,~IEEE,}
        Xingzhi Wang, and Yiu-ming~Cheung,~\IEEEmembership{IEEE Fellow}
\thanks{The work was supported in part by the National Science Foundation of China (Nos. 61673185 and  61672444), in part by the National Science Foundation of Fujian Province (Nos. 2020J01083 and 2020J01084), in part by Quanzhou City Science \& Technology Program of China (No. 2018C107R), in part by Hong Kong Baptist University (HKBU) under Grants: RC-FNRA-IG/18-19/SCI/03 and RC-IRCMs/18-19/01,  in part by the Innovation and Technology Fund of Innovation and Technology Commission of the Government of the Hong Kong under Project ITS/339/18, and in part by the SZSTC (SGDX20190816230207535). (Corresponding author:
Xin Liu.)}
\thanks{X. Liu is with the Department of Computer Science, Huaqiao University,  and also  with the Xiamen Key Laboratory of Computer Vision and Pattern Recognition \& Fujian Key Laboratory of Big Data Intelligence and Security,  Xiamen 361021, China. E-mail: xliu@hqu.edu.cn.}
\thanks{X. Wang is with the School of Electronics and Information Technology, Sun Yat-sen University, Guangzhou 510006, China. E-mail: wangxzh58@mail2.sysu.edu.cn.}
\thanks{Y.M. Cheung is with the Department of Computer Science, Hong Kong Baptist University,  Hong Kong SAR, China. E-mail: ymc@comp.hkbu.edu.hk.}
}


\markboth{IEEE Transactions on Neural Networks and Learning Systems}%
{Shell \MakeLowercase{\textit{et al.}}: Bare Demo of IEEEtran.cls for Journals}		
	
	\maketitle
	
\begin{abstract}
	Cross-modal hashing, favored for its effectiveness and efficiency, has received wide attention to facilitating efficient retrieval across different modalities. Nevertheless,
most existing methods do not sufficiently exploit the discriminative power of semantic information when learning the hash codes, while often involving time-consuming training
procedure for handling the large-scale dataset. To tackle these issues, we  formulate the learning of similarity-preserving
hash codes in terms of orthogonally rotating the semantic data so as to minimize the quantization loss of mapping
such data to hamming space, and  propose an efficient Fast  Discriminative Discrete Hashing (FDDH) approach for large-scale cross-modal retrieval. More specifically, FDDH introduces an orthogonal basis to regress the targeted hash codes  of training examples to their corresponding
 semantic labels, and  utilizes $\varepsilon$-dragging technique to provide provable large semantic margins. Accordingly, the discriminative power of semantic information can be explicitly captured  and maximized. Moreover, an orthogonal transformation scheme is further proposed to map the nonlinear embedding data into the semantic subspace, which can well guarantee the semantic
 consistency between the data feature and its semantic representation. Consequently,  an efficient closed-form solution is derived for discriminative hash code learning,
 which is very computationally efficient. In addition, an effective and stable online learning strategy is presented for optimizing modality-specific projection functions,
 featuring adaptivity to different training sizes and streaming data. The proposed  FDDH  approach theoretically approximates the bi-Lipschitz continuity, runs sufficiently fast, and
 also significantly improves the retrieval performance over the state-of-the-art methods. The source code is released at: https://github.com/starxliu/FDDH.
\end{abstract}

%
%
%

\IEEEpeerreviewmaketitle
	
	\section{Introduction}
	\IEEEPARstart{W}{ith} the explosive growth of multimedia data, automated mechanisms are needed to establish
a similarity link from one multimedia item to another if they are related to each other.
In order to maximally benefit from the richness of multi-modal media data, cross-modal retrieval has recently gained increasing attention to approximate nearest neighbors search across different modalities,  such as using image to search the relevant text documents or
	using text to search the relevant images. Nevertheless, in real multimedia searching, the multi-modal data usually
	span in different feature spaces, and each feature characterizes the
data contents from different aspects. Such modality heterogeneity
	has been widely considered as a great challenge to cross-modal
	retrieval. To alleviate this concern, early naive  studies \cite{pereira2014on} learn a common latent subspace to minimize the modality heterogeneity,
 indicating its possibility to directly compare the features from different modalities.  Although these methods have achieved impressive performance, there is still a serious limitation for them. That is, the existing subspace approaches  are computationally expensive to deal with the large-scale and high
	dimensional media data.

Hashing, favored for its low storage cost and fast retrieval speed,  has recently attracted much more attention due to its effectiveness for indexing large-scale multimedia data~\cite{TNNL2019binary}.
	The main objective of cross-modal hashing is to learn the compact binary codes for representing multiple modalities, while faithfully preserving both intra-modality similarity and
inter-modality similarity. In recent years, various cross-modal hashing researches have been devoted
to compress multi-modal data in an isomorphic Hamming space and bridge their heterogeneity gap, which can be roughly divided into  unsupervised methods \cite{Song2013Inter,Ding2014Collective,Zhou2014Latent,Long2016Composite} and supervised methods \cite{Zhang2014Large,Hong2017Cross,Mandal2017Generalized,Xu2017Learning}. Since the label information is 	helpful to construct the semantic correlations across different modalities, the supervised methods  often leverage the
semantic labels  to further improve retrieval performance over the unsupervised cases.
 In spite of some supervised methods that have achieved impressive retrieval performance, it still remains a
challenging task to achieve efficient cross-modal retrieval  mainly due to the complex integration of semantic gap, modality heterogeneity and mixed binary-integer optimization problem. For instance, the utilization of label supervision in terms of large pairwise similarity or affinity matrix~\cite{Lin2015Semantics,mandal2019generalized}  inevitably increases the computational cost during the  hash code learning process. Specifically, the discrete constraints imposed on the binary codes and hash objective functions often lead to NP-hard  optimization problems. To  simplify
such optimization, some supervised methods relax the original discrete optimization problem into the continuous one~\cite{Tang2016Supervised}, and such  relaxation scheme may deteriorate the accuracy of the learned binary codes due to the accumulated quantization error. Besides, recent supervised discrete hashing methods attempt to learn the hash code bit by bit~\cite{Xu2017Learning,XINPami}, which often involves large iterations in the learning process. In recent years, deep cross-modal hashing approaches \cite{Jiang2017Deep,Yang2018Shared},
 integrate the feature learning and  hashing code learning together, which always yield outstanding performance on 	many benchmarks. Nevertheless, these deep works often involve exhaustive search for learning optimum parameters, which is quite time-consuming.

 To the best of our knowledge, it can be well found that  existing supervised methods just consider the semantic-preserving property provided by the label supervision, which do not carefully explore the discriminative power of semantic information when learning the hash codes. Consequently, the learnt hash codes are not discriminative enough for high retrieval performance. Besides, exiting methods rarely consider to  shorten the optimization iterations during the training process,  making it unscalable to large-scale datasets. Therefore, it is still desirable to study a fast and discriminative cross-modal hashing method from a practical viewpoint.

%

In this paper, we present a Fast Discriminative Discrete Hashing (FDDH) approach to facilitating efficient retrieval across different modalities. The main contributions of the proposed FDDH approach are four-fold:
\begin{itemize}
\item The efficient learning of similarity-preserving
hash codes is newly formulated in terms of orthogonally rotating the semantic data, whereby the quantization loss of
mapping such data to hamming space can be well minimized.
\item The label values are reasonably relaxed to increase the discrimination
power of semantic information, and $\varepsilon$-dragging methodology is introduced to provide a large margin property for discriminative hash code learning. To the best of our knowledge,
this strategy has yet to be studied thus far in cross-modal hashing.
\item A novel orthogonal regression method is proposed for learning semantic-preserving
hash codes, and an efficient closed form solution is derived in a discrete and discriminative manner, which accelerate the learning process and makes less computational effort.
\item Extensive experiments on three public benchmarks highlight the advantages of FDDH under various cross-modal
retrieval scenarios and show its improved retrieval performance over the state-of-the-art ones.
\end{itemize}

The remainder of this paper is structured as follows: Section~\ref{relatedworks} briefly surveys the related works of cross-modal hashing, and Section~\ref{proposedmethod} elaborates  FDDH and its theory analysis. The experimental results and discussions are provided in Section~\ref{results}. Finally, we draw a conclusion in Section~\ref{conclusion}.
	
	\section{Related Work}\label{relatedworks}

The primary issue of cross-modal retrieval lies that the  features of different
modalities often span in different feature spaces, indicating its impossibility to be compared directly. To tackle this problem, Canonical Correlation Analysis (CCA)  \cite{Rasiwasia2010A}
is probably the most popular method that aims to learn a common latent subspace from two modalities, where the features of different modalities can be directly correlated and compared. Similarly, Partial Least Square (PLS) \cite{Li2011Cross} and Bilinear model (BLM) \cite{Tenenbaum2014Separating} also learn a
common latent subspace for cross-modal retrieval. Remarkably, these  methods do not utilize the semantic labels for
discriminative analysis. Therefore, some extensions leverage
the valuable label information to improve the retrieval
performance. For instance, the multi-view
CCA framework~\cite{gong2014a} directly links the image and text views under the semantic class labels, and other extensions, \emph{e.g.},  cluster-CCA \cite{Rasiwasia2014Cluster} and multi-label CCA \cite{Ranjan2015Multi}, have also been developed
to address cross-modal retrieval problem. Besides,  multi-modal deep models, \emph{e.g.},  multi-modal auto-encoder \cite{Ngiam2009Multimodal} and  deep CCA \cite{Andrew2013Deep},
have been recently proposed to construct more powerful subspace in
the hidden layers of neural network, while capturing the nonlinear correlation between the heterogeneous modalities. Remarkably, these deep methods  are computationally expensive to process  the large-scale and high-dimensional media data.


Cross-modal hashing  has received wide attention due to its effectiveness in reducing
memory cost and  improving query speed, and its main difficulty is to learn compact hash codes that have an additional property to preserve the semantic relationship between different
modalities. It is noted that the recent multi-modal hashing \cite{wang2015learning,zhu2020flexible} or multi-view hashing \cite{zhu2020TMM} works often fuse the features from different modalities or views to learn the comprehensive hash codes, which are essentially different from cross-modal hashing that concentrates on discovering the shared
hash codes to correlate multiple modalities. Further, multi-modal hashing is designed for multimedia search when multi-modal features are all provided at the query stage, while cross-modal hashing aims to retrieve the most relevant objects represented by other modalities for a given query characterized by one modality. In the following, we mainly survey the cross-modal hashing works.  In the past, various cross-modal hashing attempts have been proposed,
mostly in either unsupervised manner where the labels are unavailable, or supervised
manner where the labels are explicitly provided.
Unsupervised cross-modal hashing methods mainly learn the projection functions to map the original feature spaces into hamming spaces.
Accordingly, Inter-media Hashing (IMH) \cite{Song2013Inter} obtains a common hamming space by preserving the inter-view and intra-view consistency,
while Collective Matrix Factorization Hashing (CMFH) \cite{Ding2014Collective,ding2016large} jointly learns the unified hash codes and hash functions
by collective matrix factorization.  Similarly, Latent Semantic Sparse Hashing (LSSH) \cite{Zhou2014Latent} first
utilizes sparse coding and matrix factorization to extract latent semantic features, and then quantizes such latent semantic feature  for hash code generation.
 Although these methods are able to capture the semantic correlations between heterogeneous modalities, the available class label information remains unexplored and the
derived hash codes are not discriminative enough for high retrieval performance.

Supervised cross-modal hashing methods primarily exploit the available label information to learn the compact hash codes,
which can well mitigate the semantic gap between heterogeneous modalities and generally show the improved performance than that of unsupervised ones.
For instance, Semantic Correlation Maximization (SCM) \cite{Zhang2014Large} seamlessly integrates the
semantic labels into hash code learning procedures, while Semantic Preserved Hashing (SePH) \cite{Lin2015Semantics} and its extension \cite{SePH2} generate the unified binary code by modeling an affinity matrix in a probability distribution. Besides,  Supervised Matrix Factorization Hashing (SMFH) \cite{Tang2016Supervised} 	utilizes the label supervision to produce unified hash codes, while maintaining the label consistency and local geometric consistency. It is noted that hashing is essentially
a discrete learning problem, and these methods utilize relaxation-based continuous schemes to simplify the original binary
optimization problem. Nevertheless, the approximated solutions with relaxation are suboptimal, which often degrade the discriminative power of the
final hash codes, possibly due to the accumulated quantization error. In contrast to this, discrete methods try to directly solve the discrete
problem without continuous relaxation. Along this way, Discrete Cross-modal Hashing (DCH) \cite{Xu2017Learning}, Discrete Latent Factor
Model Hashing (DLFH) \cite{8636536}, Sequential Discrete Hashing \cite{liu2017sequential}, Cross-Modal Discrete Hashing (CMDH) \cite{liong2018cross-modal},  Asymmetric Discrete Cross-Modal  Hashing (ADCH) \cite{ADCMH} and Nonlinear Robust Discrete Hashing (NRDH) \cite{NRDH} directly update hash codes while retaining the discrete constraints
for more compact hash codes. In addition, Generalized Semantic Preserving Hashing (GSePH) \cite{mandal2019generalized} constructs an affinity
matrix by label supervision to discretely approximate hash codes, while Scalable Discrete Matrix Factorization Hashing (SCRATCH) \cite{8691805},
Subspace Relation Learning for Cross-modal Hashing (SRLCH) ~\cite{shen2020exploiting} and scalaBle Asymmetric discreTe Cross-modal Hashing (BATCH)~\cite{BATCH} improve the collective matrix factorization to discretely learn the hash codes. Although these supervised methods are able to achieve efficient cross-modal retrieval, they do not fully exploit the discriminative power of semantic information when learning hash codes, and  often involve a bit large iterations in the training procedures.

In recent years,  multi-modal deep learning  has  proven to be effectively in capturing the high-level correlation in different modalities. Accordingly, recent deep cross-modal hashing works \cite{Yang2018Shared,TNNL2019,Minoradd2,Minoradd1} jointly learn the high-level features and hash code in an integrated way, whereby the hash codes can be optimized
with feature representation learning through the multi-layer neural networks. Although these deep methods have shown outstanding performance on many benchmarks, they are always constrained by computational complexity and exhaustive search for learning optimum network parameters. Another potential limitation is that these deep methods still employ the binary
quantization functions to generate hash codes from the feature space, which cannot guarantee the learned binary codes to be semantically discriminative for characterizing the heterogeneous modalities.  Therefore,
it is still desirable to study the fast and discriminative indexing techniques for efficient cross-modal retrieval practically.

\section{Fast Discriminative Discrete Hashing}\label{proposedmethod}

Without loss of generality, this section mainly focuses on fast  discriminative discrete hashing  with only two modalities (\emph{i.e.}, image and text),
and the proposed cross-modal hashing framework can be easily extended to three or more modalities.

	\subsection{Notation and Problem Formulation}
		Throughout this paper, the uppercase bold font characters are utilized to denote matrices, while the lowercase bold font characters
are select to represent data vectors.
For simplicity, let ${\mathbf{X}^{t}{=}\{ \mathbf{x}_i^{t}\}_{i = 1}^n, t= 1,2}$ be the  training image-text examples,
where $\mathbf{x}_i^{t}\in \mathbb{R}^{d_t}$ is the $i$-th sample, ${d_t}$ is the feature length  in $t$-th modality, and $n$ is the training number. Without loss of generality, the data points are
assumed to be zero-centered which means $\sum\nolimits_{i = 1}^n \mathbf{x}_i^{t}{=}0$. In practice,
the zero–one matrix $\mathbf{Y}{=}[\mathbf{y}_1,\cdots,\mathbf{y}_n]\in \mathbb{R}^{c{\times}n}$ is utilized to represent the label matrix corresponding to the training samples,
where each column vector $\mathbf{y}_i{\in}\mathbb{R}^{c{\times}1}$ is simply defined as follows: if the $t$-th training sample comes from the ${j}$-th class (in general each sample belongs
to no less than one class), then the ${j}$-th element of such column vector is 1 while the remaining elements are 0.
The goal of cross-modal hashing is to learn binary codes matrix $\mathbf{H}{=}\{ {\mathbf{h}_i}\} _{i = 1}^n \in \mathbb{R}^{q{\times}n}$ for all training instances,
and modality-specific projection matrix $\mathbf{P}_t$  for  linking the original  feature space and the common hamming space, where $\mathbf{h}_i{\in}{\{-1,1\} ^{q{\times}1}}$
is $q$ bits hash code of the $i$-th instance.

	\subsection{The Proposed FDDH Methodology}\label{proposedmethod}

	\begin{figure}[tbp]
		\centering
		\includegraphics[width=3.2in]{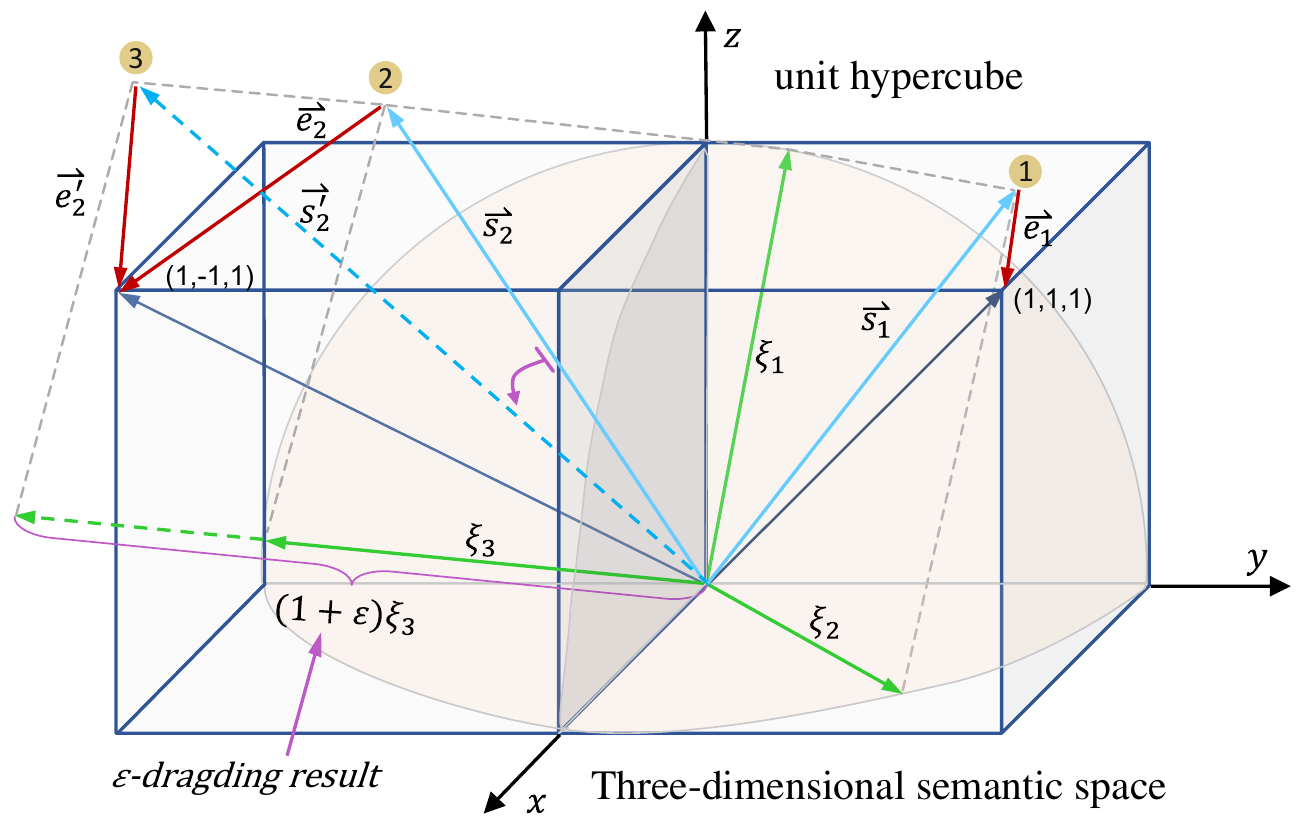}
		\caption{A geometric example of three-dimensional semantic subspace illustration for semantic-preserving and  quantization error reduction.}
		\label{fig:figure1}
	\end{figure}

%


	\subsubsection{\textbf{Semantic-preserving Hash Code Learning}}
For hashing representation learning, compactness is a critical criterion to guarantee its performance in efficient similarity
search.	Therefore, it is imperative to produce an efficient code in which the variance of
	each bit is maximized and the bits are pairwise uncorrelated. To learn the compact hash code, the orthogonal transformation is popular for discriminative hash code learning and Wang et al.~\cite{wang2015learning} present an orthogonal learning structure to reduce the redundant information lying in the hash representation. Note that, this work is designed to learn modality-specific hash codes jointly for multi-modal data representation, while cross-modal hashing learns common hash codes to preserve the semantic relationship across different modalities. Benefit from the observations of work~\cite{GongPami}, the learning of similarity-preserving binary codes can be successfully formulated in terms of orthogonally rotating zero-centered PCA-projected data,  so as to decompose correlations among hash bits and minimize the quantization error of mapping that data to the vertices of a zero-centered binary hypercube.
In practice, the length of hash bits is often larger than the number of semantic categories, i.e., $q{\geq}c$.
Geometrically, it is not difficult to find that hamming space is consistent with the vertices of unit hypercube, and we heuristically  introduce an orthogonal basis
$\mathbf{C}{=}\{ \mathbf{\xi} _i\} _{i = 1}^c{\in}\mathbb{R}^{q{\times}c}$  to learn the semantic representation.
More specifically, we propose to orthogonally rotate the
label vector $\mathbf{y}_i$ to approximate its semantic data $\mathbf{s}_i$ (\emph{i.e.}, $\mathbf{s}_i{=}\mathbf{C}\mathbf{y}_i$), while ensuring  $\mathbf{s}_i$ to be as
close as possible to the vertices of unit hypercube. Accordingly, the  optimal orthogonal basis $\mathbf{C}$ can be obtained by   minimizing the following quantization loss:
	\begin{equation}
	\mathop {\min }\limits_\mathbf{C} \sum\limits_{i = 1}^n {{{\left\| {{\mathop{\rm sgn}} (\mathbf{C}\mathbf{y}_i) - \mathbf{C}\mathbf{y}_i} \right\|}^2_2}}, ~~s.t.~\mathbf{C}^{\rm T}\mathbf{C} = \mathbf{I}_c,
	\label{eq1}
	\end{equation}
where $\mathbf{I}_c$ is a $c$-order identity matrix. A geometric example for  three-dimensional semantic subspace illustration is shown in Fig.~\ref{fig:figure1},
where $\vec{s_1}$ and $\vec{s_2}$ are semantic vectors of instances that formed by an orthogonal  basis $\mathbf{C}{=}\{ {\xi _1},{\xi _2},{\xi _3}\}$  and label vectors ${\mathbf{y}_1}{=}{(1,1,0)^{\rm T}}$, ${\mathbf{y}_2} = {(1,0,1)^{\rm T}}$. The quantization errors are attainted by computing the sum of the length of $\vec{e_1}$ and $\vec{e_2}$. As indicated in Eq.~(\ref{eq1}), the smaller quantization loss indicates the better binary code, which can well preserve the structure property of the semantic data. As hash codes are often obtained by quantizing the semantic data, Eq.~\eqref{eq1} can be rewritten as:
	\begin{equation}
	\mathop {\min }\limits_\mathbf{C} \sum\limits_{i = 1}^n {{{\left\| {\mathbf{h}_i - \mathbf{C}\mathbf{y}_i} \right\|}^2_2}}, ~~~~s.t.~\mathbf{C}^{\rm T}\mathbf{C} = \mathbf{I}_c.
	\label{eq1C}
	\end{equation}

Therefore, the learning of semantic-preserving binary codes can be formulated in terms of orthogonally rotating the semantic data to minimize the quantization loss:
\begin{equation}
	\mathop {\min }\limits_{\mathbf{H,C}} \left\| {\mathbf{H} - \mathbf{CY}} \right\|_F^2,~s.t.~\mathbf{H}{\in}{\{- 1,1\}^{q{\times}n}},~\mathbf{C}^{\rm T}\mathbf{C} = \mathbf{I}_c.
	\label{eq2}
	\end{equation}


Remarkably, the optimization problem in Eq.~\eqref{eq1C} ensures that  $\mathbf{C{y}}_i{\rightarrow}\mathbf{h}_i$. Considering the linear structural equation $\mathbf{A}\mathbf{x}{=}\mathbf{b}$, the solution of $\mathbf{x}$ can be determined when $rank(\mathbf{A}){=}rank(\mathbf{A},\mathbf{b})$. Formally, we replace $\mathbf{A}$ with $\mathbf{C}$ and $\mathbf{b}$ with $\mathbf{h}_i$, and obtain  $\mathbf{C}\mathbf{x}{=}\mathbf{h}_i$. Since $\mathbf{C}$ is constructed from the orthogonal basis in semantic space, and $\mathbf{h}_i$ can be linearly represented with these  basis, resulting $rank(\mathbf{C}){=}rank(\mathbf{C},\mathbf{h}_i)$.
Since $\mathbf{C{y}}_i{\rightarrow}\mathbf{h}_i$, $\mathbf{y}_i$  is in close proximity to the solution of $\mathbf{x}$ in equation $\mathbf{C}\mathbf{x}{=}\mathbf{h}_i$, resulting $\left\|\mathbf{y}_i{-}\mathbf{x}\right\|_2{=}\left\|\mathbf{y}_i{-}\mathbf{C}^{\rm T}\mathbf{h}_i\right\|_2{\rightarrow}{0}$, and we obtain the following equivalent equation:
	\begin{equation}
	\mathop {\min }\limits_\mathbf{C} \sum\limits_{i = 1}^n {{{\left\| {\mathbf{y}_i{-}\mathbf{C}^{\rm T}\mathbf{h}_i} \right\|}^2_2}},~~~s.t.~\mathbf{C}^{\rm T}\mathbf{C} = \mathbf{I}_c.
	\label{eq1C2}
	\end{equation}

As indicated in work~\cite{Shen2015CVPR}, the hash code can also be regressed to  its corresponding label,  Eq.~\eqref{eq1C2} is also in accordance with such feasibility. By combing all the instances,  Eq.~\eqref{eq1C2} can be rewritten in the matrix representation:
	\begin{equation}
	\mathop {\min }\limits_{\mathbf{H,C}} \left\| {\mathbf{Y} - \mathbf{C}^{\rm T}\mathbf{H}} \right\|_F^2,~s.t.~\mathbf{H}{\in}{\{ - 1,1\} ^{q{\times}n}},~\mathbf{C}^{\rm T}\mathbf{C} = \mathbf{I}_c.
	\label{eq3}
	\end{equation}

The above formulation is a typical regression problem, which regress $\mathbf{H}$ to $\mathbf{Y}$. That is, the zero–one class label vectors stipulate a
type of binary regression with target `1' for positive class and target `0' for the negative classes. Evidently, for the rigid zero–one label matrix $\mathbf{Y}$,
the Euclidean distances of regression responses between samples from different classes are a constant value, \emph{i.e.}, $\sqrt{2}$ for single label data. This is contrary to the expectation that the samples from different classes should be as far as possible.
To alleviate this problem, we propose to utilize $\varepsilon$-dragging technique to force
the regression targets of different classes moving along opposite
directions, whereby the margin between different classes can be enlarged.
That is, with a positive slack variable $\varepsilon_i$, we hope the output will
become $1{+}\varepsilon_i$ for the sample grouped into `1'  and $-\varepsilon_i$ for
the sample categorized into `0'. As shown in  Fig.~\ref{fig:figure1}, the label  $\mathbf{y}_2= (1,0,1)^{\rm T}$ is further relaxed as $\mathbf{y}_2 =(1,0,1 + \varepsilon )^{\rm T}$, and the semantic vector $\vec{s_2}$ is updated to $\vec{s'_2}$. Accordingly, the resulted length of $\vec{e'_2}$  is smaller than the original one
of $\vec{e_2}$,  and the total quantization error is reduced by $\left| {\vec{e_2}} \right| - | {\vec{e'_2}}| $.

\begin{table}[!t]
\caption{Impacts of $\varepsilon$-dragging on class label vectors.}\setlength{\tabcolsep}{0.23cm}
\begin{center}
\begin{tabular}{|c|c|c|c|}
\hline
    samples  &    label     $  \mathbf{y}^{\rm T}$ & $\mathbf{y}^{\rm T}$ after $\varepsilon$-dragging & constraint \\
\hline
       $\mathbf{x}_1$ &     $[1,0,0]$     &  $[1+\varepsilon_{11}, -\varepsilon_{12}, -\varepsilon_{13}]$   &  $\{\varepsilon_{11}, \varepsilon_{12}, \varepsilon_{13}\}{\geq}0 $     \\
\hline
         $\mathbf{x}_2$ &   $[1,0,0]$     &  $[1+\varepsilon_{21}, -\varepsilon_{22}, -\varepsilon_{23}]$   &  $\{\varepsilon_{21}, \varepsilon_{22}, \varepsilon_{23}\}{\geq}0 $    \\
\hline
         $\mathbf{x}_3$ &    $[0,1,0]$    &  $[-\varepsilon_{31}, 1+\varepsilon_{32}, -\varepsilon_{33}]$   &  $\{\varepsilon_{31}, \varepsilon_{32}, \varepsilon_{33}\}{\geq}0 $   \\
\hline
         $\mathbf{x}_4$ &     $[0,1,0]$    &    $[-\varepsilon_{41}, 1+\varepsilon_{42}, -\varepsilon_{43}]$ & $\{\varepsilon_{41}, \varepsilon_{42}, \varepsilon_{43}\}{\geq}0 $     \\
\hline
         $\mathbf{x}_5$ &   $[0,0,1]$    &    $[-\varepsilon_{51}, -\varepsilon_{52}, 1+\varepsilon_{53}]$ & $\{\varepsilon_{51}, \varepsilon_{52}, \varepsilon_{53}\}{\geq}0 $     \\
\hline
         $\mathbf{x}_6$ &    $[0,0,1]$    &    $[-\varepsilon_{61}, -\varepsilon_{62}, 1+\varepsilon_{63}]$ & $\{\varepsilon_{61}, \varepsilon_{62}, \varepsilon_{63}\}{\geq}0 $     \\
\hline
\end{tabular}
\end{center}
\label{dragging}
\end{table}

Further interpretation of  $\varepsilon$-dragging motivation is shown in Table~\ref{dragging}, which reports six single label data points in three classes and their one-hot class label vectors are listed in the second column. It can be observed that  $\mathbf{x}_1$  and  $\mathbf{x}_2$  are marked within the same class, while $\mathbf{x}_3$, $\mathbf{x}_4$, $\mathbf{x}_5$ and $\mathbf{x}_6$ are categorized into other different classes. Specifically, if  the first components of the class label vectors are gathered, we can
get values  ``1, 1, 0, 0, 0, 0'', and their values  will be relaxed into ``$1+\varepsilon_{11}$,$1+\varepsilon_{21}$,  $-\varepsilon_{31}$, $-\varepsilon_{41}$, $-\varepsilon_{51}$, $-\varepsilon_{61}$. As all $\varepsilon$ values are nonnegative, such $\varepsilon$-dragging technique could help to enlarge the distance
between different classes in case where the data points are mapped. It is noted that the real datasets may have large volume of data points and involve multiple semantic labels, and the proposed $\varepsilon$-dragging operation can be well utilized in these datasets as well. To develop a unique compact model, we utilize a constant matrix $\mathbf{B}{\in}\mathbb{R}^{c{\times}n}$ to characterize the dragging
direction, in which the $i$-th row and $j$-th column element $\mathbf{B}_{ij}$ is defined as:
	\begin{equation}
{\mathbf{B}_{ij}} = \left\{ \begin{gathered}
   + 1,{\kern 1pt} {\kern 1pt} {\kern 1pt} {\kern 1pt} if{\kern 1pt} {\kern 1pt} {\mathbf{y}_{ij}} = 1 \hfill \\
   - 1,{\kern 1pt} {\kern 1pt} {\kern 1pt} {\kern 1pt} {\kern 1pt} {\kern 1pt} otherwise \hfill \\
\end{gathered}  \right.,
	\end{equation}
where`$+ 1$' represents the positive direction and `$- 1$' means the negative direction. Performing the above $\varepsilon$-dragging operation on each
element of $\mathbf{Y}$ and recording these $\varepsilon$ values by matrix $\mathbf{E}{=}\{\varepsilon_{ij}\geq0\}\in \mathbb{R}^{c{\times}n}$,
 Eq.~(\ref{eq3}) can be rewritten the as following optimization problem:
	\begin{equation}
	\begin{array}{l}
	\mathop {\min }\limits_{\mathbf{Y,C,E}} \left\| \mathbf{Y}+\mathbf{B}\odot\mathbf{E} - \mathbf{C}^{\rm T}\mathbf{H} \right\|_F^2 + \delta \left\| \mathbf{E} \right\|_F^2\\
	~s.t.~\mathbf{H}\in {\{  - 1,1\} ^{q \times n}},~~\mathbf{C}^{\rm T}\mathbf{C} = \mathbf{I}_c,
	\end{array}
	\label{eq4}
	\end{equation}
where $\odot$ is a Hadamard product operator of matrices and $\delta$ is the weight coefficient to control the degree of relaxation. In contrast to Eq.~(\ref{eq3}), we add an  $\varepsilon$-dragging term  $\mathbf{B}\odot\mathbf{E}$ in Eq.~(\ref{eq4}) to enlarge the distances between
different classes. In this way, each sample can be regressed efficiently with a large margin between the true and false classes.
Accordingly, the learning model is converted to be an equivalently constrained optimization problem. For ease of representation,
let $\mathbf{\bar{Y}}{=}\mathbf{Y}{+}\mathbf{B}\odot\mathbf{E}$,  it can be easily found that the term $\mathop {\min }\limits_{\mathbf{E}}\left\| \mathbf{E} \right\|_F^2$ is equivalent to $\mathop {\min }\limits_{\mathbf{\bar{Y}}}\left\| \mathbf{\bar{Y}} \right\|_F^2$. Therefore, Eq.~\eqref{eq3} can be further rewritten as:
\begin{equation}
\begin{split}
\mathop {\min }\limits_{\mathbf{H,C,\bar{Y}}} &\left\| {\mathbf{\bar{Y}} - \mathbf{C}^{\rm T}\mathbf{H}} \right\|_F^2+\delta \left\| \mathbf{\bar{Y}} \right\|_F^2,\\
s.t.\;~\mathbf{H}&{\in}{\{- 1,1\}^{q{\times}n}},~\mathbf{C}^{\rm T}\mathbf{C} = \mathbf{I}_c.
	\end{split}
\label{eqdragfinal}
	\end{equation}

\subsubsection{\textbf{Semantic Embedding Learning}}

On the one hand, any suitable embedding learning algorithms, linear or nonlinear, can be utilized for mapping the data into the semantic space. In general, the semantic correlation
of multiple modalities often exists in the high-level space and the mapping functions from the raw feature space to the high-level space are highly nonlinear~\cite{Lin2015Semantics,Kulis2010Kernelized}. On the other hand, for hash representation learning, it is necessary to produce an efficient code in which the
variance of each bit is maximized and the bits are pairwise uncorrelated. To integrate these issues, we propose a simpler idea of orthogonally transforming the
data and utilize following simple yet powerful nonlinear form:
\begin{equation}
	\mathop {\min }\limits_{{{\mathbf{R}}_t}} \left\| {\mathbf{S}} - {\mathbf{R}_t^{\rm{T}}} \phi({{{\mathbf{X}}^{t}}) } \right\|_F^2,~~~s.t. ~\mathbf{R}_t^{\rm{T}}{{\mathbf{R}}_t}  =  {{\mathbf{I}}_{q}},
	\label{eqR}	
	\end{equation}
where $\mathbf{S}{\in}\mathbb{R}^{q{\times}n}$ represents the semantic data, and $\phi ( \cdot )$ is the RBF kernel \cite{Kulis2010Kernelized} which
could better capture the underlying nonlinear property in feature space. Since the semantic data is approximated by orthogonally rotating label vector,  the semantic
subspace can be further approximated by  $\mathbf{S}=\mathbf{C\bar{Y}}$. Similar to the relationship between Eq.~\eqref{eq2} and Eq.~\eqref{eq3}, Eq.~\eqref{eqR} can be transformed into following  equivalent form.
\begin{equation}
	\mathop {\min }\limits_{{{\mathbf{R}}_t}} \left\| {\phi({{{\mathbf{X}}^{t}})  - {\mathbf{R}_t}\mathbf{C\bar{Y}}}} \right\|_F^2,~~~s.t. ~\mathbf{R}_t^{\rm{T}}{{\mathbf{R}}_t} =  {{\mathbf{I}}_{q}}.
	\label{eq5}
	\end{equation}

\subsubsection{\textbf{Overall Objective Function}}
According to the orthogonal relationship, the first item $\mathop {\min }\limits_{\mathbf{H,C}} \left\| {\mathbf{\bar{Y}}{-}\mathbf{C}^{\rm T}\mathbf{H}} \right\|_F^2$ in
Eq.~\eqref{eqdragfinal} is also equivalent to $\mathop {\min }\limits_{\mathbf{H,C}} \left\| {\mathbf{H}{-}\mathbf{C}\mathbf{\bar{Y}}} \right\|_F^2$.  By integrating the  semantic-preserving learning and semantic embedding learning, the process of learning the discriminative  hash codes can be conducted by  minimizing the following objective function:
\begin{equation}
	\begin{split}
\mathop {\min }\limits_{\mathbf{H},\mathbf{R}_1,\mathbf{R}_2,\mathbf{C,\bar{Y}}} & \left\| {\mathbf{H}{-}\mathbf{C}\mathbf{\bar{Y}}} \right\|_F^2+\mu \left\| {\phi {\rm{(}}{{\mathbf{X}}^1}){-}{{\mathbf{R}}_1}{\mathbf{C\bar{Y}}}} \right\|_F^2\\&+\theta \left\| {\phi {\rm{(}}{{\mathbf{X}}^2}){-}{{\mathbf{R}}_2}{\mathbf{C\bar{Y}}}} \right\|_F^2{+}\delta \left\| \mathbf{\bar{Y}} \right\|_F^2\\
s.t.\;\; & \mathbf{C}^{\rm T}\mathbf{C} = \mathbf{I}_c, ~~~\mathbf{H}\in {\{  - 1,1\} ^{q \times n}},\\
& 	{\mathbf{R}}_1^{\rm{T}}{{\mathbf{R}}_1}={{\mathbf{I}}_{q}},~~~{\mathbf{R}}_2^{\rm{T}}{{\mathbf{R}}_2} =  {{\mathbf{I}}_{q}}.
	\end{split}
	\label{eq6}
	\end{equation}

\subsection{Discrete Optimization for FDDH}\label{optimization}

The discrete constraints imposed in Eq.~(\ref{eq6}) lead to
mixed integer optimization problems, which are generally
NP-hard. In the literature, some hashing methods discard discrete constraints and
solve a relaxed problem to simplify the optimization steps. Nevertheless,
this relaxation strategy may accumulate large quantization
error during the hash code learning process. To solve Eq.~(\ref{eq6}), the discrete optimization method is selected. For all matrix
variables $\mathbf{H}$, $\mathbf{R}_1$, $\mathbf{R}_2$, $\mathbf{C}$ and $\mathbf{\bar{Y}}$, it is convex
with respect to any single matrix variable while fixing the other ones, and an alternating optimization technique can be adopted
to iteratively solve such optimization problem until the convergence is reached. The details of the discrete optimization steps are elaborated as follows:

\textbf{Update $\mathbf{C}$}: removing the items that are irrelevant to $\mathbf{C}$ and fixing $\mathbf{H}$, $\mathbf{R}_1$, $\mathbf{R}_2$ and $\mathbf{\bar{Y}}$. Then, the sub-optimization problem derived in Eq.~(\ref{eq6}) is simplified as:
	\begin{equation}
	\begin{aligned}
\mathop {\min }\limits_{\mathbf{C}} & \left\| {\mathbf{H}{-}\mathbf{C}\mathbf{\bar{Y}}} \right\|_F^2+\mu \left\| {\phi{\rm{(}}{{\mathbf{X}}^1}){-}{{\mathbf{R}}_1}{\mathbf{C\bar{Y}}}} \right\|_F^2\\
& +\theta \left\| {\phi {\rm{(}}{{\mathbf{X}}^2}){-}{{\mathbf{R}}_2}{\mathbf{C\bar{Y}}}} \right\|_F^2\\
&  s.t.\;\; \mathbf{C}^{\rm T}\mathbf{C} = \mathbf{I}_c.
	\end{aligned}
	\label{eqC}
	\end{equation}
By expanding each item and removing the irrelevant ones, we can rewrite Eq.~\eqref{eqC} as follows:
	\begin{equation}
	\begin{aligned}
	\mathop {\min }\limits_{\mathbf{C}}& {\left\|\mathbf{C}\mathbf{\bar{Y}}\right\|_F^2{+}\mu\left\|{\mathbf{R}}_1{\mathbf{C\bar{Y}}}\right\|_F^2{+}\theta \left\| {\mathbf{R}}_2{\mathbf{C\bar{Y}}}\right\|_F^2}\\
	 &-2\text{Tr}\left(\mathbf{\bar{Y}}\left(\mathbf{H}^{\rm T}{+}\mu \phi ({{\mathbf{X}}^{1}})^{\rm T}{{\mathbf{R}}_1}{+}\theta \phi({{\mathbf{X}}^{2}})^{\rm T}{{\mathbf{R}}_2}\right)\mathbf{C}\right)\\
&  s.t.\;\; \mathbf{C}^{\rm T}\mathbf{C} = \mathbf{I}_c,
	\end{aligned}
	\label{eqC1}
	\end{equation}
where $\text{Tr}(\cdot)$ is the trace norm. Since  $\mathbf{C}^{\rm T}\mathbf{C}{=}\mathbf{I}_c$, ${\mathbf{R}}_1^{\rm{T}}{{\mathbf{R}}_1}{= }{\mathbf{I}}_{q}$ and
${\mathbf{R}}_2^{\rm{T}}{{\mathbf{R}}_2}{= }{\mathbf{I}}_{q}$, it can be easily obtained that $\left\|\mathbf{C}\mathbf{\bar{Y}}\right\|_F^2{=}\left\|\mathbf{\bar{Y}}\right\|_F^2$,
$\left\|{\mathbf{R}}_1{\mathbf{C\bar{Y}}}\right\|_F^2{=}\left\|\mathbf{\bar{Y}}\right\|_F^2$, and $\left\|{\mathbf{R}}_2{\mathbf{C\bar{Y}}}\right\|_F^2{=}\left\|\mathbf{\bar{Y}}\right\|_F^2$, and all these values are constant.
Therefore, the optimization problem in Eq.~\eqref{eqC1} is equal to maximize the following trace function:
 	\begin{equation}
	\begin{aligned}
	\mathop {\max}\limits_{\mathbf{C}}&~\text{Tr}\left(\mathbf{\bar{Y}}\left(\mathbf{H}^{\rm T}{+}\mu \phi ({{\mathbf{X}}^{1}})^{\rm T}{{\mathbf{R}}_1}{+}\theta \phi({{\mathbf{X}}^{2}})^{\rm T}{{\mathbf{R}}_2}\right)\mathbf{C}\right)\\
& ~s.t.\;\; \mathbf{C}^{\rm T}\mathbf{C} = \mathbf{I}_c.
	\end{aligned}
	\label{eqC2}
	\end{equation}

It is noted that the problem in Eq.~\eqref{eqC2} corresponds to the classic Orthogonal
Procrustes problem~\cite{schonemann1966a}, which can be approximated by Singular Value Decomposition (SVD).
For simplicity, let $\mathbf{Q}{=}\mathbf{\bar{Y}}(\mathbf{H}^{\rm T}{+}\mu \phi ({{\mathbf{X}}^1})^{\rm T}{{\mathbf{R}}_1}{+}\theta \phi({{\mathbf{X}}^2})^{\rm T}{{\mathbf{R}}_2})$,
we utilize  SVD to decompose $\mathbf{Q}$, \emph{i.e.}, $\mathbf{Q}{\approx}\mathbf{U}\mathbf{\Sigma}\mathbf{V}^{{\rm T}}$,
where $\mathbf{\Sigma}{=} diag(\sigma_1,\sigma_2,\cdots,\mathbf{\sigma}_r)$, $r{\leq}\min(q,c)$, $\mathbf{U}{\in}\mathbb{R}^{c{\times}r}$ and $\mathbf{V}{\in}\mathbb{R}^{q{\times}r}$ are the transformation matrices. Let $\mathbf{Z}{=}{{\mathbf{V}}^{\rm{T}}}{\mathbf{CU}}$,  the following properties are obtained:
	\begin{equation}
	\begin{aligned}
	\text{Tr}({\mathbf{QC)}}  &= \text{Tr}(\mathbf{U\Sigma} {\mathbf{V}^{{\rm T}}}{\mathbf{C}}) = \text{Tr}({\mathbf{V}^{{\rm T}}}{\mathbf{C}}\mathbf{U}\mathbf{\Sigma})\\
	& = \text{Tr}(\mathbf{{Z}\Sigma}) = \sum\limits_{i = 1}^r {{z_{ii}}{\sigma _i}} \le \sum\limits_{i = 1}^r {{\sigma _i}}.
	\end{aligned}
	\label{eqC3}
	\end{equation}

Evidently, the upper bound in Eq.~\eqref{eqC3} can be achieved if $\mathbf{Z}{=}\mathbf{I}_r$, whereby the optimal solution of $\mathbf{C}$ can be obtained by:
	\begin{equation}
	{\mathbf{C}=}{{\mathbf{V}}}\mathbf{U}^{\rm{T}}.
	\label{eqC4}
	\end{equation}

In general, the length of hash code  is often larger than the number of semantic categories, \emph{i.e.}, $q{\geq}{c}$, and it is reasonable to set $r{=}c$  during the learning process.


\textbf{Update $\mathbf{R_1},\mathbf{R_2}$}: removing the items that are irrelevant to $\mathbf{R_1},\mathbf{R_2}$ and fixing $\mathbf{C}, \mathbf{\bar{Y}}$, Eq.~\eqref{eq6} can be further simplified  as:
	\begin{equation}
	\mathop {\min }\limits_{{\mathbf{R}_1}}\left\| {\phi(\mathbf{X}^{1}){-}{\mathbf{R}_1}\mathbf{CY}} \right\|_F^2,~~s.t.~{\mathbf{R}}_1^{\rm{T}}{{\mathbf{R}}_1} =  {{\mathbf{I}}_{q}},
	\label{eqR1}
	\end{equation}
	\begin{equation}
	\mathop {\min }\limits_{{\mathbf{R}_2}}\left\| {\phi(\mathbf{X}^{2}){-}{\mathbf{R}_2}\mathbf{CY}} \right\|_F^2,~~s.t.~{\mathbf{R}}_2^{\rm{T}}{{\mathbf{R}}_2} =  {{\mathbf{I}}_{q}}.
	\label{eqR2}
	\end{equation}
	
Equivalently, the optimization problems in Eqs.~\eqref{eqR1} and \eqref{eqR2}  are equal to maximize the following trace functions:
	\begin{equation}
	\mathop {\max }\limits_{{\mathbf{R}_1}}~\text{Tr}\left(\mathbf{C\bar{Y}}\phi ({{\mathbf{X}}^{1}})^{\rm T} {\mathbf{R}_1} \right),~~s.t.~{\mathbf{R}}_1^{\rm{T}}{{\mathbf{R}}_1} =  {{\mathbf{I}}_{q}},
	\label{eqR11}
	\end{equation}
	\begin{equation}
	\mathop {\max }\limits_{{\mathbf{R}_2}}~\text{Tr}\left(\mathbf{C\bar{Y}}\phi ({{\mathbf{X}}^{2}})^{\rm T} {\mathbf{R}_2} \right),~~s.t.~{\mathbf{R}}_2^{\rm{T}}{{\mathbf{R}}_2} =  {{\mathbf{I}}_{q}}.
	\label{eqR22}
	\end{equation}

Similarly, the optimal solutions of $\mathbf{R}_1$ and $\mathbf{R}_2$  can be approximated by respectively computing the SVD of  $\mathbf{C\bar{Y}}\phi ({{\mathbf{X}}^{1}})^{\rm T}$ and $\mathbf{C\bar{Y}}\phi ({{\mathbf{X}}^{2}})^{\rm T}$, \emph{i.e.}, $\mathbf{C\bar{Y}}\phi ({{\mathbf{X}}^{1}})^{\rm T}{\approx} \mathbf{U}_1\mathbf{\Sigma}_1\mathbf{V}_1^{\rm T}$ and $\mathbf{C\bar{Y}}\phi ({{\mathbf{X}}^{2}})^{\rm T}{\approx}\mathbf{U}_2\mathbf{\Sigma}_2\mathbf{V}_2^{\rm T}$. By referring to Eqs.~\eqref{eqC3} and~\eqref{eqC4}, the solution of $\mathbf{R}_1$ and $\mathbf{R}_2$ can be achieved by:
	\begin{equation}
	\mathbf{R}_1 = {\mathbf{V}}_1\mathbf{U}^{\rm T}_1,~~\mathbf{R}_2 = {\mathbf{V}}_2\mathbf{U}^{\rm T}_2.
	\label{eqRU}
	\end{equation}

\textbf{Update $\mathbf{\bar{Y}}$}: removing the items that are irrelevant to $\mathbf{\bar{Y}}$ and fixing  $\mathbf{C}, \mathbf{H}, \mathbf{R}_1, \mathbf{R}_2$, Eq.~(\ref{eq6}) can be
rewritten as:
		\begin{equation}
	\begin{aligned}
	\mathop {\min }\limits_{\mathbf{\bar{Y}}} &\left(1{+}\mu{+}\theta{+}\delta \right)\text{Tr}\left({{\mathbf{\bar{Y}}}^{\rm{T}}}{\mathbf{\bar{Y}}}\right) - 2\text{Tr}\left({{\mathbf{H}}^{\rm{T}}}{\mathbf{C\bar{Y}}}\right)\\
	&{-}2\mu\text{Tr}\left(\phi {({\mathbf{X}^1})^{\rm{T}}}{{\mathbf{R}}_1}{\mathbf{C\bar{Y}}}\right){\rm{-}}2\theta\text{Tr}\left(\phi {({\mathbf{X}^2})^{\rm{T}}}{{\mathbf{R}}_2}{\mathbf{C\bar{Y}}}\right).
	\end{aligned}
	\label{eqY}
	\end{equation}

Further, the subproblem of Eq.~\eqref{eqY} can be simplified as:
	\begin{equation}
	\mathop {\min }\limits_{\mathbf{\bar{Y}}}(1{+}\mu{+}\theta{+}\delta)\text{Tr}\left({{\mathbf{\bar{Y}}}^{\rm{T}}}{\mathbf{\bar{Y}}}\right){-}2\text{Tr}\left(\mathbf{W}^{\rm{T}}\mathbf{\bar{Y}}\right),
	\label{eqY1}
	\end{equation}
where $\mathbf{W}{=}\mathbf{C}^{\rm T}\left(\mathbf{H}{+}\mu{{\mathbf{R}}_1^{\rm{T}}}\phi{({\mathbf{X}^1})}{\rm{+}}\theta{{\mathbf{R}}_2^{\rm{T}}}\phi{({\mathbf{X}^2})}\right)$.
To solve such minimization problem, the gradient descent method is selected. More specifically, the derivative
of all the terms in Eq.~\eqref{eqY1} with respect to $\mathbf{\bar{Y}}$ is  derived, and its optimal value is attained at $\mathbf{\bar{Y}'}{=}\frac{1}{1{+}\mu{+}\theta{+}\delta}\mathbf{W}$. It is noted that the original label matrix $\mathbf{Y}$ is now extended to be $\mathbf{\bar{Y}}{=}\mathbf{Y}{+}\mathbf{B}\odot\mathbf{E}$, where $\mathbf{E}$ is the  $\varepsilon$-dragging matrix that utilized to enlarge the distances between different classes. Evidently, the updating result for the false class should be smaller than zero, and the
regression result for the true class should be larger than one. Therefore, the updating scheme of optimized target matrix $\mathbf{\bar{Y}}$
 is further regularized as:
	\begin{equation}
	{\mathbf{\bar{Y}}_{ij}} = \left\{ {\begin{array}{*{20}{c}}
		{\min(\mathbf{\bar{Y}'}_{ij},~0),{\rm{~~if}}~\mathbf{y}_{ij}={\rm{0}}} \\
		{\max(\mathbf{\bar{Y}'}_{ij},~1),{\rm{~~if}}~\mathbf{y}_{ij}={\rm{1}}}
		\end{array}} \right..
	\label{eqYU}
	\end{equation}


\textbf{Update $\mathbf{H}$}: removing the items that are irrelevant to $\mathbf{H}$ and fixing $\mathbf{C}, \mathbf{\bar{Y}}$,  Eq.~\eqref{eq6} can be simplified as:
	\begin{equation}
\mathop {\min }\limits_{\mathbf{H}} \left\| {\mathbf{H}{-}\mathbf{C}\mathbf{\bar{Y}}} \right\|_F^2, ~~s.t.\;\;\mathbf{H}\in {\{- 1,1\} ^{q \times n}}.
	\label{eq12}
	\end{equation}

 The discrete solution of $\mathbf{H}$ can be computed from the embedding  matrix $\mathbf{C}\mathbf{\bar{Y}}$ and  an efficient close-form solution is approximated by  thresholding such data matrix as:
	\begin{equation}
	\mathbf{H} = {\mathop{\rm sgn}} (\mathbf{C\bar{Y}})
	\label{eqHH}
	\end{equation}

It is noted that the proposed FDDH approach has a closed-form solution for hash code learning and only requires a single step to obtain all bits, which is highly efficient in comparison with  bit by bit learning scheme~\cite{Xu2017Learning}. The main procedures of the proposed FDDH method are summarized in Algorithm~\ref{alg:1}.

	\begin{algorithm}[t]
		\renewcommand{\algorithmicrequire}{\textbf{Input:}}
		\renewcommand{\algorithmicensure}{\textbf{Output:}}
		\caption{Learning Algorithm for FDDH}
		\label{alg:1}
		\begin{algorithmic}[1]
			\REQUIRE Training data $\mathbf{X}^{1}$, $\mathbf{X}^{2}$; code length $q$; semantic labels $\mathbf{L}$, parameters $\mu$, $\theta$, $\delta$ and $\gamma$.
			\STATE Initialize $\mathbf{C}, \mathbf{R}_1,\mathbf{R}_2$ as random matrix respectively.
			\STATE Initialize $\mathbf{\bar{Y}}$ = $\mathbf{Y}$ and $\mathbf{H}\in {\{  - 1,1\} ^{q \times n}}$ randomly.
\REPEAT
			\STATE Update $\mathbf{C}$ via Eq~(\ref{eqC4});
			\STATE Update $\mathbf{R}_1$ and $\mathbf{R}_2$ via Eq.~(\ref{eqRU});
			\STATE Update $\mathbf{H}$ via Eq~(\ref{eq12});
			\STATE Update $\mathbf{Y}$ via Eq~(\ref{eqYU});
			\UNTIL{convergency or reaching maximum iterations.}
			\ENSURE Obtain hash code matrix $\mathbf{H}$ via Eq~(\ref{eqHH}).
		\end{algorithmic}
	\end{algorithm}

	\subsection{Out-of-Sample Extension}
The hash function is designed to project high-dimensional real-value
features to low-dimensional binary space. For any unseen query sample, it is straightforward to predict its hash codes via the modality-specific hash function. On the one hand,
the offline learning method is the standard way to learn such modality-specific hashing projections, which keep unchanged for all new coming data. Similar to most existing methods,
the offline learning strategy is configured within the proposed FDDH framework. On the other hand, the multimedia data points often
continuously arrive in a streaming fashion. If the training data is increasingly
accumulated, the offline learning method needs to recalculate the hash functions on the whole database, which is computationally
inefficient. Therefore,  it is particularly important to develop an efficient online learning strategy to deal with the new query data. In the following,
we elaborate the offline strategy and the newly proposed online strategy in tandem.

	\subsubsection{\textbf{Offline Strategy}}
	As introduced in Section~\ref{proposedmethod}, the discriminative hash codes can be well obtained by optimizing the  objective function in Eq.~(\ref{eq6}).
Since the linear hash function  cannot characterize the nonlinearity embedded in real-world data, we refer to the training process and first utilize the RBF kernel
to capture the underlying nonlinear information in feature space. Then, the  modality-specific projection ${\mathbf{P}}_t$ can be obtained by minimizing the following formulation:
	\begin{equation}
	{G_{{\rm{offline}}}({{\mathbf{P}}_t}{}}){=}\left\| {{\mathbf{H}}{-}{{\mathbf{P}}_t}\phi ({{\mathbf{X}}^{t}})} \right\|_F^2{+}\gamma \left\|{{{\mathbf{P}}_t}} \right\|_F^2,~t{=}1,2
	\label{eq19}
	\end{equation}
	where $\gamma$ is the hyper-parameter for regularization term. The solution of $\mathbf{P}_t$  can be computed by a regularized
linear regression method and its optimal solution is obtained when the gradient of Eq.~(\ref{eq19}) is equal to zero:
	\begin{equation}
	{\mathbf{P}_t} = \mathbf{H}\phi {(\mathbf{X}^{t})^{\rm T}}{(\phi (\mathbf{X}^{t})\phi {(\mathbf{X}^{t})^{\rm T}}{+}{\gamma\mathbf{I}})^{-1}},~t{=}1,2
	\label{eq20}
	\end{equation}
	
	For any query sample $\mathbf{x}^{t}_i$ in $t$-th modality, the corresponding hash code $\mathbf{h}^{t}_i$ can be directly generated by:
	\begin{equation}
	\mathbf{h}^{t}_i = {\mathop{\rm sgn}} (\mathbf{P}_t\phi (\mathbf{x}^{t}_i)).
	\label{eq21}
	\end{equation}
	

	\subsubsection{\textbf{Online Strategy}}\label{online}
In practice, the multimedia data often comes in a streaming fashion, and the hashing projection functions derived from the offline strategy keep unchanged for all new data points. If the new data is increasingly accumulated, the offline learning method may accumulate large
quantization error during the hash code  learning process. To alleviate this concern, we  present a simple but effective online learning strategy to
adaptively learn the modality-specific projections, by minimizing the following formulation:
	\begin{equation}
	G_{\rm{online}}({{\mathbf{P}}_t}) = {G_{\rm{offline}}}({{\mathbf{P}}_t}){+}\left\| \mathbf{H}_s^{t}{-}{{\mathbf{P}}_t}\phi ({\mathbf{X}}_s^{t}) \right\|_F^2,
	\label{eq22}
	\end{equation}	
	where $\mathbf{H}_s^t$ is the corresponding hash code matrix of the stream data ${\mathbf{X}}_s^{t}$. The optimal solution of $\mathbf{P}_t$ and $\mathbf{H}_s^t$  can be obtained
 when the gradients of Eq.~\eqref{eq22} are equal to zeros. Accordingly, the modality-specific projection $\mathbf{P}_t$ can be computed by:
	\begin{equation}
	\begin{aligned}	
	{{\mathbf{P}}_t} = &({\mathbf{H}}\phi {({{\mathbf{X}}^{t}})^{\rm{T}}}{\rm{ + }}{{\mathbf{H}}_s^{t}}\phi {({\mathbf{X}}_s^{t})^{\rm{T}}})\\
	&{(\phi ({{\mathbf{X}}^{t}})\phi {({{\mathbf{X}}^{t}})^{\rm{T}}}{+}\phi ({\mathbf{X}}_s^{t})\phi {({\mathbf{X}}_s^{t})^{\rm{T}}}{+}{\gamma\mathbf{I}})^{-1}}
	\end{aligned}
	\label{eq23}
	\end{equation}	

It is noted that the computation results of  ${\mathbf{H}}\phi ({\mathbf{X}}^{t})^{\rm{T}}$ and $\phi ({\mathbf{X}}^{t})\phi ({\mathbf{X}}^{t})^{\rm{T}}$ can be stored as constants during the learning process.
For the $t$-th modality of stream data $\mathbf{X}_s^{t}$, the  corresponding hash codes $\mathbf{H}_s^{t}$ can be obtained by:
	\begin{equation}
	{{\mathbf{H}}_s^{t}} = {\mathop{\rm sgn}} ({{\mathbf{P}}_t}\phi ({\mathbf{X}}_s^{t})),~t=1,2
	\label{eq24}	
	\end{equation}		

The final solutions of $\mathbf{P}_t$ and $\mathbf{H}_s^{t}$ are obtained  iteratively by repeating  Eqs.~(\ref{eq23}) and~(\ref{eq24}) until the procedure converges.


	\subsection{Theoretical Analysis}\label{theory}
	The proposed FDDH approach aims to produce discriminative semantic-preserving hash codes in a fast way, while reducing the quantization error.
This section shows the theoretical analysis to prove its effectiveness. In addition, the theoretical analysis of online strategy is also given to prove its stability.

	1)~\emph{\textbf{Efficiency of Semantic-preserving}}: To facilitate analysis of the relationship between the semantic information and hash code, we utilize single data point  for illustration.
 According to Eqs.~\eqref{eq1C} and \eqref{eq1C2}, we can obtain the following relationship:
	\begin{equation}		
	\begin{array}{l}
	\mathbf{h}_i = \mathbf{C{y}}_i + \mathbf{e}_i;~~ \mathbf{{y}}_i = \mathbf{C}^{\rm T}\mathbf{h}_i + \mathbf{e}'_i,
	\end{array}
	\label{eq25}
	\end{equation}
where $\mathbf{{y}}_i$ denotes the label value for the $i$-th sample, $\mathbf{e}_i$ and $\mathbf{e}'_i$ denote the quantization error and  regression error, respectively.
Accordingly, given two label vectors $\mathbf{y}_i$ and $\mathbf{y}_j$ from the semantic space, and their corresponding hash representations $\mathbf{h}_i$ and $\mathbf{h}_j$, we can obtain the following equations:
	\begin{equation}	
	\begin{aligned}
	\left\| {\mathbf{h}_i-\mathbf{h}_j} \right\|_2 & = \left\| {\mathbf{C}(\mathbf{y}_i - \mathbf{y}_j)} {+}\left( \mathbf{e}_i-\mathbf{e}_j\right)\right\|_2,\\
\| {\mathbf{y}_i-\mathbf{y}_j}- (& \mathbf{e}'_i - \mathbf{e}'_j) \|_2 = \| {\mathbf{C}^{\rm T}(\mathbf{h}_i-\mathbf{h}_j)}  \|_2.\\
		\end{aligned}
	\label{eq26}
	\end{equation}

For any two vectors $\mathbf{u},\mathbf{v}$, there are two norm properties $\left\| {\mathbf{uv}} \right\|_2 \le \left\| \mathbf{u} \right\|_2\left\| \mathbf{v} \right\|_2$ and
$\left\| {\mathbf{u}{+}\mathbf{v}} \right\|_2{\le}\left\| \mathbf{u} \right\|_2{+}\left\| \mathbf{v} \right\|_2$, and the similar properties can also  be  found in matrix representation.
According to the definition of Frobenius norm, we can further obtain that $\left\| {\mathbf{C}}^{\rm T}\right\|_F{=}\left\| {\mathbf{C}} \right\|_F{=}\kappa$, where $\kappa$ is a constant. According to the norm property, we can find  $\left\| \mathbf{C}\right\|_2{\leq}\left\| \mathbf{C}\right\|_F$, and obtain the following semantic structure preservation property:
	\begin{equation}	
	\begin{array}{l}
\frac{1}{\kappa}\left\| {\mathbf{y}_i{-}\mathbf{y}_j} \right\|_2{-}{\epsilon_2{(i,j)}}{\ll}\left\| {\mathbf{h}_i{-}\mathbf{h}_j} \right\|_2{\ll}\kappa \left\| {\mathbf{y}_i{-} \mathbf{y}_j} \right\|_2{+}{{{\epsilon_1}}(i,j)},
	\end{array}
	\label{eq27}
	\end{equation}
where ${\epsilon}_1(i,j){=}\left\| {\mathbf{e}_i - \mathbf{e}_j} \right\|_2$ and ${{\epsilon_2}(i,j)} = \frac{1}{\kappa}\left\| {\mathbf{e}'_i - \mathbf{e}'_j} \right\|_2$.
Remarkably, if the bound error terms ${\epsilon}_1$ and ${\epsilon}_2$ are removed, it is not difficult to find that the relationship in Eq.~(\ref{eq27}) is  in accordance with
bi-Lipschitz continuity. Equivalently, if the quantization error $(\mathbf{e}_i,\mathbf{e}_j)$ and the regression error $(\mathbf{e}'_i,\mathbf{e}'_j)$  are reduced perfectly,
the error terms  ${\epsilon}_1$ and ${\epsilon}_2$ shall closely equate to 0, whereby  Eq. \eqref{eq27} approximates the relationship of bi-Lipschitz continuity.
That is,  the smaller error bounds guarantee that $\mathbf{h}_i$ is similar to $\mathbf{h}_j$ when $\mathbf{y}_i$ is similar to $\mathbf{y}_j$. Remarkably, minimizing the objective function in  Eqs.~\eqref{eq2} and  \eqref{eq3}  can well reduce  $\mathbf{e}_i$ and $\mathbf{e}'_i$  because they
aim to find $\mathbf{h}_i$ with small quantization error and  regression error. Therefore, the semantic  consistency between the semantic information and hash code are well preserved within FDDH framework.

\begin{figure}[tbp]
\centering
   \includegraphics[width=9cm]{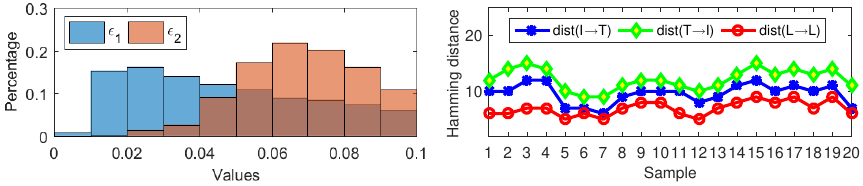}
   \vspace{-0.6cm}
   \caption{Left: The histogram distribution of bound errors. Right: Hamming distances for semantic preserving illustration.}
\label{fig:figure2}
\end{figure}

Next, we investigate the value distributions concerning to the error terms. Specifically,  5000 instances are randomly selected from the MIRFlickr dataset to compute $\mathbf{e}_i$ and $\mathbf{e}'_i$ by Eq.~(\ref{eq25}). In order to eliminate the influence of magnitude of input data pairs, we refer to the magnitude of $\left\| {\mathbf{y}_i - \mathbf{y}_j} \right\|_2$ and respectively compute the normalized relative errors of ${{{\left|{\epsilon_1}(i,j)\right|}} \mathord{\left/
	{\vphantom {{{\epsilon_1}(i,j)} {\left\| {\mathbf{y}_i - \mathbf{y}_j} \right\|}}} \right.
	\kern-\nulldelimiterspace} {\kappa\left\| {\mathbf{y}_i - \mathbf{y}_j} \right\|_2}}$  and $\kappa{\left|{\epsilon_2}(i,j)\right| \mathord{\left/
	{\vphantom {\mathbf{{\epsilon_2}(i,j)} {\left\| {\mathbf{y}_i - \mathbf{y}_j} \right\|}}} \right.
	\kern-\nulldelimiterspace} {\left\| {\mathbf{y}_i - \mathbf{y}_j} \right\|_2}}$, which can be efficiently utilized to show the impacts of the error terms resulted by Eq.~\eqref{eq27}.
As shown in the left part of Fig.~\ref{fig:figure2}, we draw the value histograms of error terms and record their number proportions among the 25M data pairs.
It can be observed that all the data errors ${\epsilon}_1$ and ${\epsilon}_2$ fall into the small range $[0,0.1]$, which means that these errors terms have very little impacts
to the right side of Eq.~(\ref{eq27}). In addition, we compute the Hamming distances (32 bits) between one randomly selected instance and  other 20 different data instances.
As shown in the right part of Fig.~\ref{fig:figure2}, it can be clearly observed that the Hamming distances of Image to Text (I${\rightarrow}$T), Text to Image (T${\rightarrow}$I)
and Label to Label (L${\rightarrow}$L) often have similar tendency, which indicate that the hash codes derived  from the proposed FDDH framework can well hold
the semantic-preserving property between heterogeneous modalities.



	\begin{table*}[!t]
		\footnotesize
		\centering
		\caption{Quantitative comparisons of mAP and top-50 Precision on PASCAL-VOC-2007, and the best results are highlighted in bold.}\setlength{\tabcolsep}{2.03mm}
		\begin{tabular}{|c|c|c|c|c|c|c|c|c|c|c|c|c|c|}
			\hline
			\multicolumn{2}{|c|}{\multirow{2}[6]{*}{Method/Dataset}} & \multicolumn{6}{c|}{Handcrafted Features}     & \multicolumn{6}{c|}{CNN Visual Features} \\
			\cline{3-14}    \multicolumn{2}{|c|}{} & \multicolumn{3}{c|}{mAP} & \multicolumn{3}{c|}{Top-50 precision} & \multicolumn{3}{c|}{mAP} & \multicolumn{3}{c|}{Top-50 precision} \\
			\cline{3-14}    \multicolumn{2}{|c|}{} & 32 bits & 64 bits & 128 bits & 32 bits & 64 bits & 128 bits & 32 bits & 64 bits & 128 bits & 32 bits & 64 bits & 128 bits \\
			\hline
			\multirow{10}[2]{*}{I${\rightarrow}$T} & CMFH~\cite{ding2016large}  & 0.3337 & 0.3270 & 0.3244 & 0.4309 & 0.4340 & 0.4388 & 0.4406 & 0.4645 & 0.4799 & 0.6390 & 0.6664 & 0.6913 \\
			& CCQ~\cite{Long2016Composite}   & 0.3100 & 0.2987 & 0.3062 & 0.3997 & 0.3663 & 0.4062 & 0.3760 & 0.3899 & 0.4194 & 0.4910 & 0.5160 & 0.5685 \\
			& IMH~\cite{Song2013Inter}   & 0.2888 & 0.2759 & 0.2653 & 0.4021 & 0.3875 & 0.3697 & 0.3543 & 0.3174 & 0.2948 & 0.5505 & 0.5209 & 0.4921 \\
			& SePH\_km \cite{Lin2015Semantics} & 0.5274 & 0.5375 & 0.5480 & 0.5683 & 0.5820 & 0.5992 & 0.7233 & 0.7265 & 0.7348 & 0.8215 & 0.8305 & 0.8382 \\
			& GSePH\_km \cite{Mandal2017Generalized}& 0.5468 & 0.5612 & 0.5672 & 0.5973 & 0.6136 & 0.6107 & 0.7582 & 0.7746 & 0.7841 & 0.8404 & 0.8466 & 0.8499 \\
			& DCH~\cite{Xu2017Learning}   & 0.4287 & 0.4300 & 0.4238 & 0.4684 & 0.4720 & 0.4780 & 0.4120 & 0.4032 & 0.3874 & 0.4688 & 0.4661 & 0.4741 \\
			& DLFH~\cite{8636536}  & 0.4433 & 0.4710 & 0.4858 & 0.5223 & 0.5519 & 0.5662 & 0.4330 & 0.4707 & 0.4927 & 0.5072 & 0.5575 & 0.5865 \\
			& SCRATCH~\cite{8691805}  & 0.4869 & 0.4984 & 0.5000 & 0.5442 & 0.5508 & 0.5424 & 0.7092 & 0.7097 & 0.7067 & 0.8168 & 0.8157 & 0.8210 \\
            & SRLCH~\cite{shen2020exploiting} &     0.4827 &     0.4776 &     0.4966 &     0.5994 &    0.5977 &     0.6212 &    0.6892 &   0.6802 &     0.7229 &     0.8440 &     0.8433 &     0.8607 \\
           & BATCH~\cite{BATCH} &   0.5530 &  0.5640 &  0.5801 &   0.6108 &    0.6203 &   0.6256 &    0.7450 &   0.7576 &    0.7726 &   0.8302 &  0.8536 &     0.8588 \\
			& DCMH~\cite{Jiang2017Deep}  & \textit{--} & \textit{--} & \textit{--} & \textit{--} & \textit{--} & \textit{--}  & 0.5248 & 0.5514 & 0.5899 & 0.6408 & 0.6645 & 0.7298 \\
			& FDDH  & \textbf{0.5615} & \textbf{0.5728} & \textbf{0.5832} & \textbf{0.6231} & \textbf{0.6261} & \textbf{0.6306} & \textbf{0.7722} & \textbf{0.7970} & \textbf{0.8029} & \textbf{0.8508} & \textbf{0.8604} & \textbf{0.8615} \\
			\hline
			\hline
			\multirow{10}[2]{*}{T${\rightarrow}$I} & CMFH~\cite{ding2016large}  & 0.3369 & 0.3380 & 0.3364 & 0.5423 & 0.5735 & 0.5755 & 0.4142 & 0.4364 & 0.4547 & 0.6810 & 0.7194 & 0.7420 \\
			& CCQ~\cite{Long2016Composite}   & 0.2952 & 0.3123 & 0.3072 & 0.4307 & 0.4524 & 0.4681 & 0.4622 & 0.4864 & 0.5092 & 0.7214 & 0.7677 & 0.7828 \\
			& IMH~\cite{Song2013Inter}   & 0.4385 & 0.4009 & 0.3630 & 0.7406 & 0.7185 & 0.6712 & 0.4777 & 0.4010 & 0.3561 & 0.7961 & 0.7352 & 0.6700 \\
			& SePH\_km \cite{Lin2015Semantics} & 0.8077 & 0.8183 & 0.8319 & 0.9381 & 0.9479 & 0.9537 & 0.8253 & 0.8248 & 0.8369 & 0.9440 & 0.9504 & 0.9554 \\
			& GSePH\_km \cite{Mandal2017Generalized}& 0.8631 & 0.8842 & 0.8939 & 0.9534 & 0.9577 & 0.9582 & 0.8751 & 0.8928 & 0.8999 & 0.9544 & 0.9576 & 0.9583 \\
			& DCH~\cite{Xu2017Learning}   & 0.8160 & 0.8288 & 0.8213 & 0.9188 & 0.9290 & 0.9299 & 0.8094 & 0.8243 & 0.8158 & 0.9177 & 0.9250 & 0.9290 \\
			& DLFH~\cite{8636536}  & 0.7607 & 0.8175 & 0.8418 & 0.8625 & 0.9210 & 0.9444 & 0.7672 & 0.8257 & 0.8405 & 0.8726 & 0.9309 & 0.9462 \\
			& SCRATCH~\cite{8691805}  & 0.8278 & 0.8371 & 0.8233 & 0.9437 & 0.9435 & 0.9419 & 0.8309 & 0.8208 & 0.8076 & 0.9445 & 0.9417 & 0.9426 \\
     & SRLCH~\cite{shen2020exploiting} &   0.8085 &    0.7827 &     0.8274 &    0.9589 &    0.9559 &   \textbf{0.9607} &   0.8015 &   0.7840 &    0.8320 &     \textbf{0.9579} &     0.9573 &     \textbf{0.9608} \\
     &BATCH~\cite{BATCH} &   0.8569 &    0.8763 &    0.8902 &    0.9507 &   0.9569 &   0.9592 &  0.8645 &    0.8748 &   0.8881 &   0.9517 &    0.9590 &     0.9563 \\
			& DCMH~\cite{Jiang2017Deep}  & \textit{--} & \textit{--} & \textit{--} & \textit{--} & \textit{--} & \textit{--}  & 0.5477 & 0.5818 & 0.6345 & 0.6604 & 0.6938 & 0.7839 \\
			& FDDH  & \textbf{0.9048} & \textbf{0.9182} & \textbf{0.9275} & \textbf{0.9590} & \textbf{0.9585} & 0.9606 & \textbf{0.9002} & \textbf{0.9216} & \textbf{0.9258} & 0.9573 & \textbf{0.9619} & 0.9596 \\
			\hline
		\end{tabular}%
		\label{tab:addlabel1}%
	\end{table*}%

	\begin{table*}[!t]
		\footnotesize
		\centering
		\caption{Quantitative comparisons of mAP and top-50 Precision on MIRFlickr, and the best results are highlighted in bold.}\setlength{\tabcolsep}{2.03mm}
		\begin{tabular}{|c|c|c|c|c|c|c|c|c|c|c|c|c|c|}
			\hline
			\multicolumn{2}{|c|}{\multirow{2}[6]{*}{Method/Dataset}} & \multicolumn{6}{c|}{Handcrafted Features}     & \multicolumn{6}{c|}{CNN Visual Features} \\
			\cline{3-14}    \multicolumn{2}{|c|}{} & \multicolumn{3}{c|}{mAP} & \multicolumn{3}{c|}{Top-50 precision} & \multicolumn{3}{c|}{mAP} & \multicolumn{3}{c|}{Top-50 precision} \\
			\cline{3-14}    \multicolumn{2}{|c|}{} & 32 bits & 64 bits & 128 bits & 32 bits & 64 bits & 128 bits & 32 bits & 64 bits & 128 bits & 32 bits & 64 bits & 128 bits \\
			\hline
			\multirow{10}[2]{*}{I${\rightarrow}$T} & CMFH~\cite{ding2016large}  & 0.5722 & 0.5582 & 0.5581 & 0.6361 & 0.5924 & 0.5879 & 0.5650 & 0.5650 & 0.5652 & 0.5970 & 0.5973 & 0.5980 \\
			& CCQ~\cite{Long2016Composite}   & 0.5848 & 0.5850 & 0.5812 & 0.6662 & 0.6721 & 0.6635 & 0.6560 & 0.6588 & 0.6645 & 0.8049 & 0.8255 & 0.8364 \\
			& IMH~\cite{Song2013Inter}   & 0.5718 & 0.5685 & 0.5650 & 0.6383 & 0.6321 & 0.6288 & 0.6134 & 0.5994 & 0.5878 & 0.7831 & 0.7791 & 0.7647 \\
			& SePH\_km \cite{Lin2015Semantics} & 0.6679 & 0.6713 & 0.6710 & 0.7240 & 0.7344 & 0.7312 & 0.7885 & 0.7913 & 0.7949 & 0.8915 & 0.8980 & 0.9051 \\
			& GSePH\_km \cite{Mandal2017Generalized}& 0.6570 & 0.6649 & 0.6694 & 0.7204 & 0.7325 & 0.7391 & 0.7879 & 0.8002 & 0.8035 & 0.8960 & 0.9045 & 0.9073 \\
			& DCH~\cite{Xu2017Learning}   & 0.6906 & 0.7017 & 0.6974 & 0.7523 & 0.7523 & 0.7598 & 0.7710 & 0.7919 & 0.7798 & 0.8928 & 0.8942 & 0.8983 \\
			& DLFH~\cite{8636536}  & 0.7194 & 0.7284 & 0.7325 & 0.7239 & 0.7433 & 0.7478 & 0.8259 & 0.8414 & 0.8464 & 0.8836 & 0.8965 & 0.8963 \\
    		& SCRATCH~\cite{8691805}  & 0.7109 & 0.7259 & 0.7282 & 0.8039 & 0.8156 & 0.8289 & 0.8126 & 0.8259 & 0.8319 & 0.9261 & 0.9173 & 0.9190 \\
			& SRLCH~\cite{shen2020exploiting} &    0.6341 &     0.6263 &     0.6525 &   0.7572 &  0.7806 &    0.8117 &    0.7172 &   0.7367 &    0.7447 &     0.9101 &    0.9234 &     0.9183 \\
    & BATCH~\cite{BATCH} &    0.7316 &   0.7403 &   0.7511 &   0.8309 &   0.8326 &   0.8340 &    0.8345 &     0.8454 &     0.8752 &     0.9397 &     0.9486 &    0.9407 \\
	& DCMH~\cite{Jiang2017Deep}  &  \textit{--} & \textit{--} & \textit{--} & \textit{--} & \textit{--} & \textit{--}  & 0.7246 & 0.7306 & 0.7420 & 0.8355 & 0.8573 & 0.8866 \\
			& FDDH  & \textbf{0.7392} & \textbf{0.7578} & \textbf{0.7631} & \textbf{0.9074} & \textbf{0.9127} & \textbf{0.9256} & \textbf{0.8515} & \textbf{0.8741} & \textbf{0.8843} & \textbf{0.9537} & \textbf{0.9590} & \textbf{0.9607} \\
			\hline
			\hline
			\multirow{10}[2]{*}{T${\rightarrow}$I} & CMFH~\cite{ding2016large}  & 0.5718 & 0.5562 & 0.5560 & 0.6231 & 0.5891 & 0.5960 & 0.5627 & 0.5625 & 0.5626 & 0.5963 & 0.5967 & 0.5962 \\
			& CCQ~\cite{Long2016Composite}   & 0.5782 & 0.5799 & 0.5748 & 0.6248 & 0.6501 & 0.6424 & 0.6553 & 0.6556 & 0.6617 & 0.8105 & 0.8203 & 0.8333 \\
			& IMH~\cite{Song2013Inter}   & 0.5710 & 0.5685 & 0.5651 & 0.6441 & 0.6422 & 0.6350 & 0.6189 & 0.6041 & 0.5919 & 0.8093 & 0.8124 & 0.7962 \\
			& SePH\_km \cite{Lin2015Semantics} & 0.7102 & 0.7158 & 0.7206 & 0.8122 & 0.8295 & 0.8392 & 0.7533 & 0.7550 & 0.7584 & 0.8483 & 0.8586 & 0.8603 \\
			& GSePH\_km \cite{Mandal2017Generalized}& 0.7004 & 0.7100 & 0.7166 & 0.8210 & 0.8289 & 0.8405 & 0.7440 & 0.7557 & 0.7614 & 0.8450 & 0.8500 & 0.8581 \\
			& DCH~\cite{Xu2017Learning}   & 0.7760 & 0.7963 & 0.7923 & 0.8779 & 0.8787 & 0.8914 & 0.7746 & 0.7946 & 0.7864 & 0.8743 & 0.8789 & 0.8924 \\
			& DLFH~\cite{8636536}  & 0.8161 & \textbf{0.8267} & 0.8347 & 0.8659 & 0.8603 & 0.8713 & \textbf{0.8150} & \textbf{0.8292} & 0.8340 & 0.8610 & 0.8668 & 0.8608 \\
			& SCRATCH~\cite{8691805}  & 0.7694 & 0.7855 & 0.7896 & 0.8886 & 0.8982 & 0.9032 & 0.7783 & 0.7889 & 0.7939 & 0.8883 & 0.8734 & 0.8772 \\
            & SRLCH~\cite{shen2020exploiting} &    0.6747 &     0.6635 &    0.6957 &    0.8517 &     0.8545 &     0.9157 &     0.6755 &     0.6827 &    0.6908 &    0.8608 &    0.8993 &   0.8918 \\
            & BATCH~\cite{BATCH} &    \textbf{0.8037} &   0.8156 &   0.8297 &   0.9131 &    0.9080 &   0.9191 &    0.8097 &     0.8218 &     0.8304 &     0.8858 &    0.9177 &    0.9127 \\
			& DCMH~\cite{Jiang2017Deep}  & \textit{--} & \textit{--} & \textit{--} & \textit{--} & \textit{--} & \textit{--}  & 0.7660 & 0.7735 & 0.7769 & 0.8760 & 0.8879 & 0.8934 \\
			& FDDH  & 0.8022 & 0.8250 & \textbf{0.8357} & \textbf{0.9485} & \textbf{0.9521} & \textbf{0.9588} & 0.8125 & 0.8247 & \textbf{0.8402} & \textbf{0.9495} & \textbf{0.9566} & \textbf{0.9578} \\
			\hline
		\end{tabular}%
		\label{tab:addlabel2}%
	\end{table*}%
	
	
	\begin{table*}[!t]
		\footnotesize
		\centering
		\caption{Quantitative comparisons of mAP and top-50 precision on NUS-WIDE dataset, and the best results are highlighted in bold.}\setlength{\tabcolsep}{2.03mm}
		\begin{tabular}{|c|c|c|c|c|c|c|c|c|c|c|c|c|c|}
			\hline
			\multicolumn{2}{|c|}{\multirow{2}[6]{*}{Method/Dataset}} & \multicolumn{6}{c|}{Handcrafted Features}     & \multicolumn{6}{c|}{CNN Visual Features} \\
			\cline{3-14}    \multicolumn{2}{|c|}{} & \multicolumn{3}{c|}{mAP} & \multicolumn{3}{c|}{Top-50 precision} & \multicolumn{3}{c|}{mAP} & \multicolumn{3}{c|}{Top-50 precision} \\
			\cline{3-14}    \multicolumn{2}{|c|}{} & 32 bits & 64 bits & 128 bits & 32 bits & 64 bits & 128 bits & 32 bits & 64 bits & 128 bits & 32 bits & 64 bits & 128 bits \\
			\hline
			\multirow{10}[2]{*}{I${\rightarrow}$T} & CMFH~\cite{ding2016large}  & 0.3429 & 0.3433 & 0.3431 & 0.4201 & 0.4235 & 0.4240 & 0.3462 & 0.3463 & 0.3463 & 0.4159 & 0.4159 & 0.4181 \\
			& CCQ~\cite{Long2016Composite}   & 0.3867 & 0.3928 & 0.3944 & 0.5200 & 0.5283 & 0.5313 & 0.4913 & 0.4967 & 0.5062 & 0.6878 & 0.7148 & 0.7379 \\
			& IMH~\cite{Song2013Inter}   & 0.3738 & 0.3609 & 0.3548 & 0.5086 & 0.4770 & 0.4632 & 0.4347 & 0.4011 & 0.3865 & 0.6588 & 0.6660 & 0.6576 \\
			& SePH\_km \cite{Lin2015Semantics} & 0.5628 & 0.5752 & 0.5811 & 0.5908 & 0.6115 & 0.6217 & 0.7402 & 0.7511 & 0.7575 & 0.8128 & 0.8257 & 0.8340 \\
			& GSePH\_km \cite{Mandal2017Generalized}& 0.5565 & 0.5710 & 0.5769 & 0.6045 & 0.6129 & 0.6202 & 0.7445 & 0.7561 & 0.7634 & 0.8286 & 0.8333 & 0.8374 \\
			& DCH~\cite{Xu2017Learning}   & 0.6398 & 0.6443 & 0.6626 & 0.6633 & 0.6209 & 0.6515 & 0.7829 & 0.8172 & 0.8404 & 0.8609 & 0.8541 & 0.8661 \\
			& DLFH~\cite{8636536}  & 0.6636 & 0.6768 & 0.6849 & 0.7752 & 0.7697 & 0.7785 & 0.8332 & 0.8548 & 0.8624 & 0.9086 & 0.9265 & 0.9298 \\
			& SCRATCH~\cite{8691805}  & 0.6649 & 0.6674 & 0.6747 & 0.7382 & 0.7406 & 0.7695 & 0.8081 & 0.8197 & 0.8277 & 0.8872 & 0.8880 & 0.8908 \\
            & SRLCH~\cite{shen2020exploiting} &    0.6144 &    0.6202 &    0.6411 &     0.7899 &   0.8032 &     0.7848 &     0.7850 &     0.7830 &     0.8073 &   0.9123 &    0.9153 &    0.9239 \\
            &BATCH~\cite{BATCH} &    0.6820 &    0.6908 &     0.7019 &     0.8048 &    0.8026 &     0.8158 &     0.8298 &     0.8368 &     0.8432 &     0.9169 &    0.9219 &    0.9205 \\
			& DCMH~\cite{Jiang2017Deep}  & \textit{--} & \textit{--} & \textit{--} & \textit{--} & \textit{--} & \textit{--} & 0.6344 & 0.6389 & 0.6514 & 0.7270 & 0.7456 & 0.7755 \\
			& FDDH  & \textbf{0.6970} & \textbf{0.6910} & \textbf{0.7118} & \textbf{0.8315} & \textbf{0.8288} & \textbf{0.8596} & \textbf{0.8452} & \textbf{0.8578} & \textbf{0.8689} & \textbf{0.9237} & \textbf{0.9391} & \textbf{0.9359} \\
			\hline
			\hline
			\multirow{10}[2]{*}{T${\rightarrow}$I} & CMFH~\cite{ding2016large}  & 0.3418 & 0.3422 & 0.3421 & 0.4101 & 0.4116 & 0.4107 & 0.3439 & 0.3440 & 0.3442 & 0.4305 & 0.4359 & 0.4415 \\
			& CCQ~\cite{Long2016Composite}   & 0.3881 & 0.3945 & 0.3976 & 0.5345 & 0.5520 & 0.5564 & 0.5016 & 0.5087 & 0.5195 & 0.7080 & 0.7336 & 0.7415 \\
			& IMH~\cite{Song2013Inter}   & 0.3705 & 0.3605 & 0.3530 & 0.5224 & 0.5017 & 0.4698 & 0.4405 & 0.4033 & 0.3884 & 0.7152 & 0.7048 & 0.7012 \\
			& SePH\_km \cite{Lin2015Semantics} & 0.6670 & 0.6738 & 0.6705 & 0.7528 & 0.7685 & 0.7701 & 0.7045 & 0.7138 & 0.7226 & 0.7663 & 0.7760 & 0.7894 \\
			& GSePH\_km \cite{Mandal2017Generalized}& 0.6523 & 0.6700 & 0.6776 & 0.7632 & 0.7623 & 0.7792 & 0.7026 & 0.7205 & 0.7222 & 0.7732 & 0.7876 & 0.7883 \\
			& DCH~\cite{Xu2017Learning}   & 0.7822 & 0.8018 & 0.8172 & 0.8339 & 0.8185 & 0.8219 & 0.7597 & 0.7936 & 0.8121 & 0.8211 & 0.8105 & 0.8287 \\
			& DLFH~\cite{8636536}  & 0.8032 & \textbf{0.8228} & \textbf{0.8265} & 0.8927 & 0.8973 & 0.8934 & 0.7987 & \textbf{0.8217} & \textbf{0.8258} & 0.8750 & 0.8851 & 0.8855 \\
			& SCRATCH~\cite{8691805} & 0.7879 & 0.7946 & 0.8013 & 0.8520 & 0.8620 & 0.8655 & 0.7830 & 0.7928 & 0.7928 & 0.8616 & 0.8589 & 0.8490 \\
            & SRLCH~\cite{shen2020exploiting} &  0.7383 &  0.7465 &   0.7692 &   0.8827 &   0.8744 &     0.8694 &   0.7484 &     0.7504 &     0.7666 &   0.8853 &    0.8794 &    0.8732 \\
            & BATCH~\cite{BATCH} &    0.7949 &     0.8099 &     0.8105 &     0.8892 &    0.8962 &     0.9062 &     0.7945 &     0.8035 &     0.8073 &    0.8876 &    0.8897 &    0.8896 \\
		& DCMH~\cite{Jiang2017Deep}  & \textit{--} & \textit{--} & \textit{--} & \textit{--} & \textit{--} & \textit{--} & 0.6722 & 0.6769 & 0.6863 & 0.7251 & 0.7389 & 0.7609 \\
			& FDDH  & \textbf{0.8133} & 0.8111 & 0.8244 & \textbf{0.9024} & \textbf{0.9024} & \textbf{0.9234} & \textbf{0.8054} & 0.8173 & 0.8167 & \textbf{0.8923} & \textbf{0.9130} & \textbf{0.9143} \\
			\hline
		\end{tabular}%
		\label{tab:addlabel3}%
	\end{table*}%
\begin{figure*}[t]
\begin{center}
	\includegraphics[width=18.2cm]{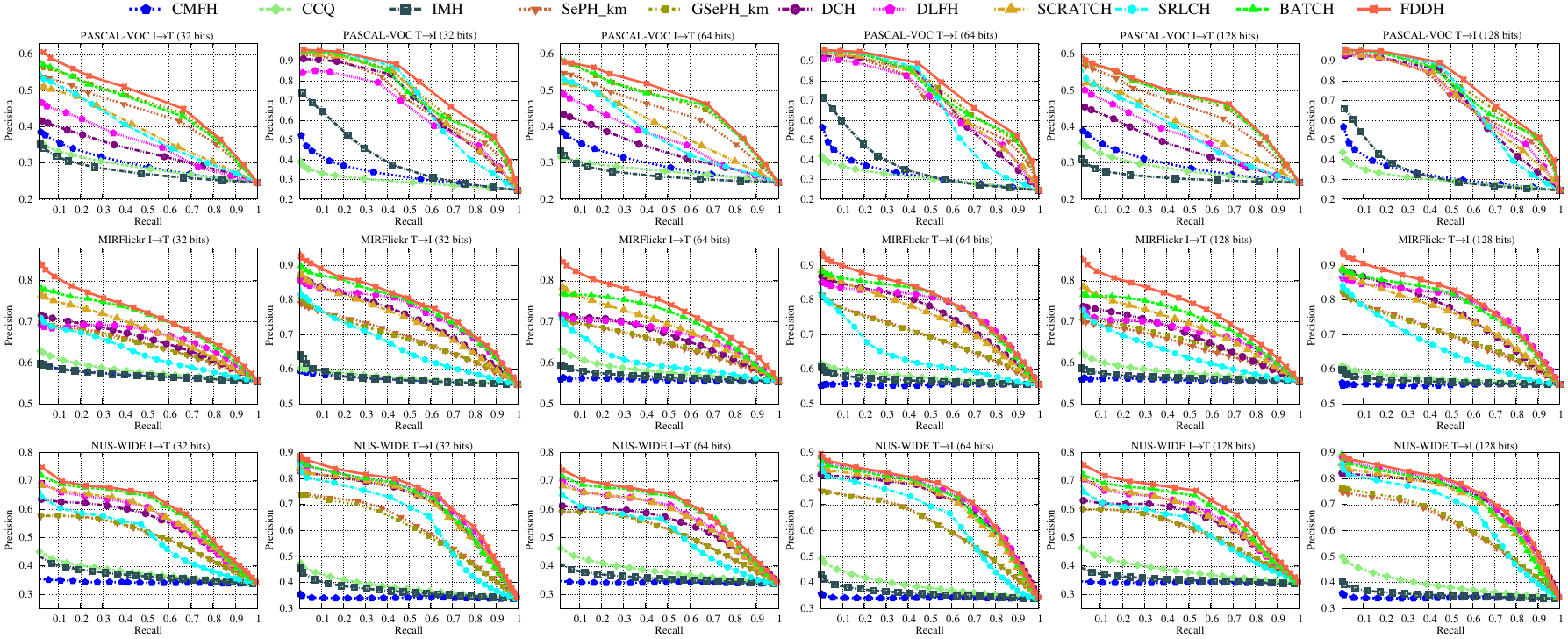}
\end{center}
\vspace{-0.2cm}
	\caption{Precision-recall curves obtained by different approaches and tested on different datasets.}
	\label{fig:prcurve}
\end{figure*}
2)~\emph{\textbf{The Stability of Online Strategy}}: We further discuss another potential benefit of the newly proposed
online hash function learning strategy, which stabilizes the hash code generation for out-of-sample extensions. In a stable algorithm, the output hash codes do not change significantly if a training example is replaced with an independent and identically distributed (iid) one. According to Eq.~\eqref{eq24}, the hash code learning for new data is closely related to the modality-specific projections. More specifically,
let ${\mathbf{X}^t_s}$ be the  new streaming dataset of $t$-th modality, and  ${\mathbf{X}}^{t/i}_s$ be the dataset with the $i$-th example in ${\mathbf{X}^t_s}$  replaced with an
iid one, the proposed online hash projection strategy holds the following stable property:
	\begin{equation}	
	 {\left\| {{\mathbf{P}_t}{(\mathbf{X}^t_s)} - {\mathbf{P}_t}{({\mathbf{X}}^{t/i}_s})} \right\|_F} \le {\mathbf{\Theta}(n)},
	\label{eq28}
	\end{equation}
where ${\mathbf{\Theta}(n)}$ converges to zero as the sample size $n$ goes to infinity. In the appendix, we provide its detailed proof.
	
3)~\emph{\textbf{Efficiency of Complexity Analysis}}: The computational complexity of FDDH mainly involves RBF mapping and the optimization in Eq. (\ref{eq6}).
For RBF mapping, whose complexity is $\mathcal{O}({m^2}{+}kdn)$, where $d{=}\max ({d_1},{d_2})$, $m$ is the number of instances selected to compute the kernel width,
and $k$ is the number of anchor points. For optimization in  Eq. (\ref{eq6}), whose complexity is $\mathcal{O}(n(q{+}c{+}qd{+}qc{+}{d^2}{+}{q^2}){+}{q^3}{+}q{d^2}{+}{d^3})$.
Since $c{\le}q{<}d{\ll}n$, the optimization complexity can be simplified as $\mathcal{O}(n(q{+}c{+}{d^2}){+}{d^3})$. Let $t$ be the iterative number to converge, the overall
complexity is approximated as ${\rm O}({m^2}{+}kdn{+}((q{+}c{+}{d^2})n{+} {d^3})t)$, which is linear to $n$ and it is very competitive to existing methods. In practice,
the iteration number is always less than 15, and more illustrations about the learning speed will be discussed in Section~\ref{results}.

\section{Experiments}\label{results}
	\subsection{Experimental Settings}
1)~\emph{\textbf{Datasets}}: The popular multi-modal PASCAL-VOC-2007~\cite{EveringhamThePascalVisual}, MIRFlickr \cite{Huiskes2010New} and NUS-WIDE \cite{Chua2009NUS} datasets
are selected for evaluation. Similar to \cite{Yang2018Shared}, PASCAL-VOC-2007 dataset (abbreviated as PASCAL-VOC) is divided into train, val, and test subsets. By dropping those pairs without text annotation, we conduct experiments on trainval and test splits, which respectively contain 5000 and 4919 pairs. For MIRFlickr, we keep 20015 image-text
pairs whose textual tags appear more than 20 times, and randomly select 2000 instances as a query set, while choosing the rest as training set.
For NUS-WIDE dataset, we select 186,577 annotated instances  from the top 10 most frequent concepts to guarantee that each concept has abundant training samples,
and randomly select 1866 instances as a query set. For these three datasets, the image is respectively represented as 512-d GIST feature vector,
512-d SIFT feature vector, and 500-d BoW feature vector, and the text   characterized by 399-d, 1386-d and 1000-d BoW feature vector. These handcrafted features are very useful for evaluating the effectiveness of different learning algorithms. In addition, the recent convolutional
neural network (CNN) has been popularized for visual feature extraction, and we also extract 4096-d CNN visual features from the last fully connected layer of classic VGG19 model \cite{Simonyan2014Very,Russakovsky2015ImageNet}.

2)~\emph{\textbf{Baseline Methods}}: For meaningful comparisons, the state-of-the-art unsupervised methods (\emph{i.e.}, CMFH \cite{ding2016large}, CCQ \cite{Long2016Composite} and IMH \cite{Song2013Inter}) and supervised methods
 (SePH \cite{Lin2015Semantics}, GSePH \cite{Mandal2017Generalized}, DCH \cite{Xu2017Learning}, DLFH \cite{8636536}, SCRATCH \cite{8691805}, SRLCH~\cite{shen2020exploiting} and BATCH~\cite{BATCH})  are selected for evaluation.
It is noted that the recent deep cross-modal
hashing methods \cite{TNNL2019,Minoradd2,Minoradd1} jointly learn the high-level feature representations and hash code in an integrated way, and the proposed framework is totally
different from those works. In that sense, it is really difficult
to perform a relatively fair and meaningful comparison with these approaches appropriately. In spite of such difference, we also select one representative deep cross-modal hashing method (DCMH \cite{Jiang2017Deep}) for comparison.
  As CMFH \cite{ding2016large}, SePH \cite{Lin2015Semantics} and GSePH \cite{Mandal2017Generalized} method are computationally
expensive in training process, it is difficult to learn their corresponding hash
functions on the whole NUS-WIDE dataset. Following the training strategy adopted in literatures \cite{Lin2015Semantics,Xu2017Learning}, we  randomly select 10000 instances from its retrieval set to learn the hash functions in the training process, and then  utilize the learned hash functions to
generate binary codes for all instances in the dataset. For all the baselines, we utilize the source codes and initialize the relevant parameters kindly provided by the respective authors.


\begin{figure*}[t]
\begin{center}
	\includegraphics[width=18.2cm]{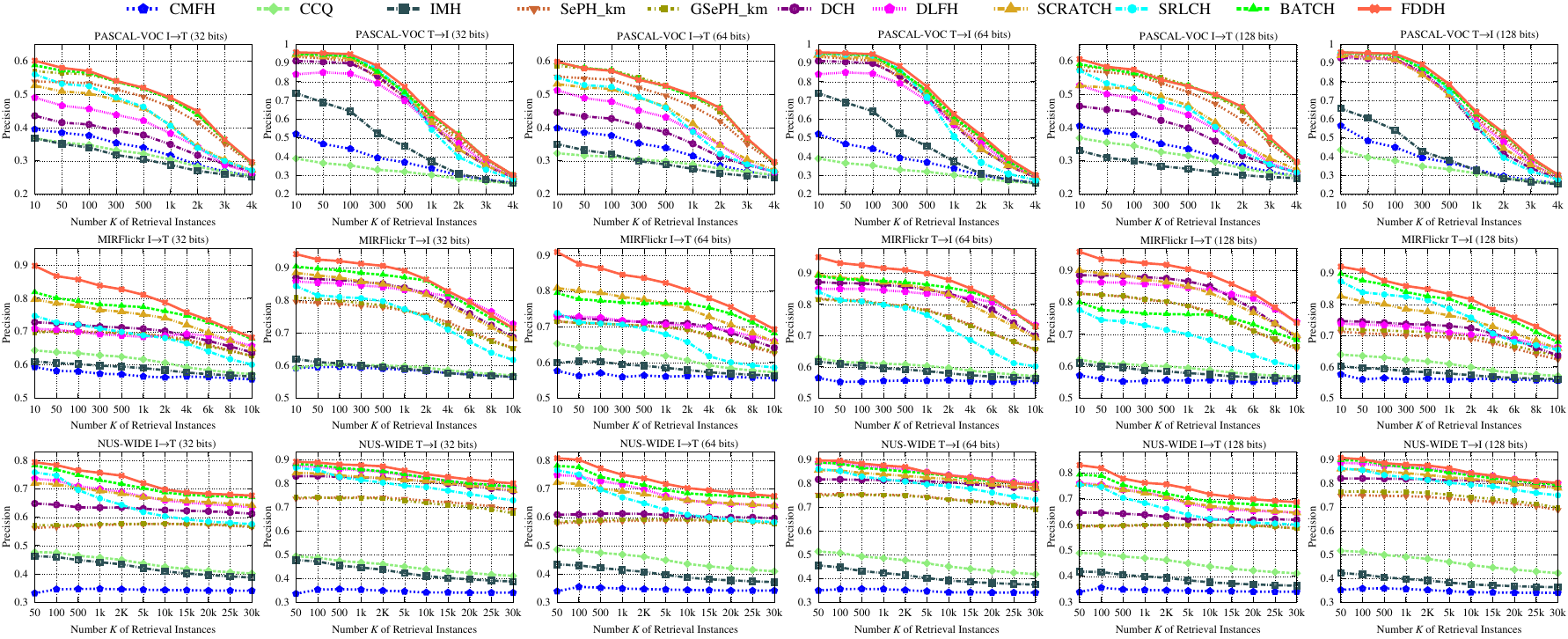}
\end{center}
\vspace{-0.2cm}
	\caption{The representative top-$K$ precision curves obtained by different approaches and tested on hand-crafted features.}
	\label{fig:figure3}
\end{figure*}

3)~\emph{\textbf{Evaluation Metric}}: The goal of  cross-modal hashing is
to index the relevant neighbors from the database of another modality, and the relevant instances are defined as those sharing as least one semantic label with the query.  Similar to most works~\cite{Mandal2017Generalized,Xu2017Learning},   mean average precision (mAP), top-$K$ precision, and  precision-recall curves are selected for quantitative analysis, including retrieving text with given image (I${\rightarrow}$T) and retrieving image with given text (T${\rightarrow}$I).
In general, the larger  mAP scores and top-$K$ precision values often indicate
the better retrieval performance.  In the experiments, parameters $\mu$, $\theta$, $\delta$ are
empirically set at $\{10^0,10^{-3},10^{3}\}$, $\{10^{-2},10^{-3},10^{3}\}$, $\{10^{-3},10^{-3},10^3\}$, respectively, for PASCAL-VOC, MIRFlickr and NUS-WIDE multi-modal datasets.

\subsection{Results and Discussion}\label{results2}
	1)~\emph{\textbf{Results of Retrieval Performances}}: The mAP scores and top-50 precision values tested with different datasets are summarized in Tables~\ref{tab:addlabel1}, ~\ref{tab:addlabel2} and~\ref{tab:addlabel3}, respectively.
For handcrafted features, it can be observed that the proposed FDDH approach  has achieved
very promising cross-modal retrieval performances in different hash length settings, and outperforms most baselines.
For instance, the proposed FDDH approach has delivered much better retrieval performances than that generated by unsupervised methods, \emph{i.e.}, CMFH, CCQ and IMH, and also yielded comparable or even the better retrieval performances than that generated by the competing supervised methods, (\emph{i.e.}, SePH\_km, GSePH\_km, DCH, DLFH, SCRATCH, SRLCH and BATCH). The main reason lies that those unsupervised methods intuitively learn the hash codes from original feature space to Hamming space, and the hash
codes learned in an unsupervised way are not discriminative enough. As a result, the corresponding semantic similarity is not well preserved in the Hamming space and the relevant retrieval performances are a bit poor.
In contrast to this, the supervised cross-modal hashing methods often deliver better retrieval performances, and the proposed FDDH method always yields the highest mAP scores in most cases and delivers the best top-50 precisions in all hash lengthes.
 For instance, the mAP scores obtained by FDDH (32 bits on T${\rightarrow}$I)
reach up to 0.9048, 0.8022 and 0.8133, respectively, evaluated on PASCAL-VOC, MIRFlickr
and NUS-WIDE datasets. Specifically, DLFH~\cite{8636536} proposes a novel discrete latent factor model to learn the binary hash codes without continuous relaxation, which performs well on some T${\rightarrow}$I retrieval task, mainly tested on MIRFlickr and NUS-WIDE datasets. Comparatively speaking, the proposed FDDH approach outperforms DLFH method in most cases. It is noted that the top-50 precision values obtained by the proposed FDDH approach are all higher than that produced by the DLFH method, which indicates that the proposed FDDH approach is able to search much more similar samples at the top ranked instances. That is, the hash codes derived from the proposed framework are more discriminative and semantically meaningful, which can well guarantee the semantic consistency between the data and its semantic representation for better cross-modal retrieval.

For the CNN Visual Features, it can  also be found that almost all competing baselines yield the improved
retrieval performances over the  handcrafted features in most cases.
Accordingly, the proposed FDDH approach often boosts the retrieval performances in different
hash length settings, and significantly outperforms most state-of-the-art baselines. For instance,  if
the hash length is set at 64 bits, the mAP scores  obtained by FDDH and tested on I${\rightarrow}$T task are higher than 0.79, 0.87, and 0.85, respectively,  evaluated on PASCAL-VOC, MIRFlickr and NUS-WIDE datasets.
Remarkably, the proposed FDDH method is designed to explicitly learn the discriminative semantic-preserving hash codes, which can achieve very competitive and even better performances compared to deep leaning method, \emph{i.e.}, DCMH. For instance, the mAP scores obtained by FDDH are higher than the results generated by DCMH in all cases. For some cases, the handcrafted features embedded within the  proposed FDDH approach yield better performance than the CNN visual features. The possible reasons are two-fold: 1) The handcrafted features associated with the semantic supervision, orthogonal regression and $\varepsilon$-dragging operation are able to produce more discriminative hash codes for better retrieval performances; 2) The mapping functions from the raw feature space to the high-level semantic space are highly nonlinear, and the handcrafted features associated with RBF mapping are able to capture the underlying nonlinear information within the visual data and therefore perform comparable with the CNN visual features.


Further, the precision-recall curves and top-$K$ precision curves tested on different feature representations are shown in Fig.~\ref{fig:prcurve} and Fig.~\ref{fig:figure3}, respectively.
On the one hand, the precision-recall curves show that the proposed FDDH approach has achieved
the comparable cross-modal retrieval performances in different hash length settings, and outperformed most baselines.
On the other hand, top-$K$ precision indicates the change of precision with respect to
the number of top-ranked $K$ instances exhibited to the users. As shown in Fig.~\ref{fig:figure3}, the top-$K$ precision curves indicate that the proposed FDDH method always yields the highest precision scores than the baselines with the number of retrieved
instances ($K$) changes, both in handcrafted visual features and CNN visual features.
This indicates that the proposed FDDH approach is able to index much more similar
samples at the beginning, which is very important for building a practical retrieval system. The main superiority contributed to these very competitive
performances lies that the hash codes derived from FDDH are more discriminative and interpretable to characterize the heterogeneous data samples,
while faithfully preserving both intra-modality similarity and
inter-modality similarity.

	2)~\emph{\textbf{Results of Ablation Studies}}: Within the proposed FDDH framework, the label relaxing  and RBF mapping schemes are carefully considered for efficient cross-modal
hashing. Next, we further evaluate the effectiveness of each
learning module and heuristically validate the performance of different
learning modules, \emph{i.e.}, FDDH without label relaxation (FDDH\_NR) and its further extension without RBF mapping (FDDH\_NRM). To be specific, their main objective formulations,  simplified directly from Eq.~\eqref{eq6}, are denoted as $\left\| {\mathbf{H}{-}\mathbf{C}\mathbf{Y}} \right\|_F^2{+}\mu \left\| {\phi {\rm{(}}{{\mathbf{X}}^1}){-}{{\mathbf{R}}_1}{\mathbf{CY}}} \right\|_F^2{+}\theta \left\| {\phi {\rm{(}}{{\mathbf{X}}^2}){-}{{\mathbf{R}}_2}{\mathbf{CY}}} \right\|_F^2$ and $\left\| {\mathbf{H}{-}\mathbf{C}\mathbf{Y}} \right\|_F^2+\mu \left\| {{{\mathbf{X}}^1}{-}{{\mathbf{R}}_1}{\mathbf{CY}}} \right\|_F^2{+}\theta \left\| { {{\mathbf{X}}^2}{-}{{\mathbf{R}}_2}{\mathbf{CY}}} \right\|_F^2$, respectively. Accordingly, the mean  mAP scores (m-mAP) and mean top-50 precision (m-top50) values, averaged on I${\rightarrow}$T and T${\rightarrow}$I tasks with all hash bits
 (\emph{i.e.}, 32, 64 and 128), are recorded to validate these different learning mechanisms.


 As illustrated in Table~\ref{tab:addlabel4}, it can be found that the m-mAP scores and m-top50 values attained by FDDH\_NR and FDDH\_NRM have also delivered very competitive
  performances. On the one hand, the reasonable relaxation of label values is able to offer a large class margin for discriminative analysis, which can promote the discriminative power of hash codes.
   On the other hand, the utilization of RBF mapping could capture the nonlinear structure of input data to improve the retrieval performance. Remarkably, the m-mAP scores obtained by FDDH
  are higher than that produced by FDDH\_NR and FDDH\_NRM in all cases, while the m-top50 values generated by FDDH yield the best retrieval precisions. That is, the integration of relaxed label value learning and RBF
  mapping could yield more discriminative hash codes and therefore significantly boost the retrieval performance.
	
	\begin{table}	
		\centering	
			\caption{Ablation studies of FDDH on VOC-PASCAL dataset.}\setlength{\tabcolsep}{0.08cm}
\label{tab:addlabel4}
		\begin{tabular}{c|c|cc|cc}
			\toprule[1pt]
				\multirow{2}{*}{Dataset} & \multirow{2}{*}{Method} & \multicolumn{2}{c|}{Handcrafted feature} & \multicolumn{2}{c}{CNN visual feature} \\
			\cline{3-6}   &       & m-mAP   & m-top50 & m-mAP   & m-top50 \\
			\hline
		\multirow{3}{*}{PASCAL-VOC}	& FDDH\_NR & 0.7440 & 0.7911 & 0.8410 & 0.9003 \\
			& FDDH\_NRM & 0.6559 & 0.7092 & 0.6506 & 0.7142 \\
			& FDDH  & \textbf{0.7447} & \textbf{0.7930} & \textbf{0.8533} & \textbf{0.9086} \\
\hline
          \multirow{3}{*}{MIRFlickr}  &FDDH\_NR & 0.7463 & 0.8693 & 0.8110 & 0.9212 \\
			&FDDH\_NRM & 0.7396 & 0.8535 & 0.8047 & 0.9100 \\
			&FDDH  & \textbf{0.7872} & \textbf{0.9342} & \textbf{0.8479} & \textbf{0.9562} \\
\hline
	 \multirow{3}{*}{NUS-WIDE} & FDDH\_NR & 0.7521 & 0.8698 & \textbf{0.8364} & 0.9186 \\
			& FDDH\_NRM & 0.7378 & 0.7663 & 0.8267 & 0.8676 \\
			& FDDH  & \textbf{0.7581} & \textbf{0.8747} & 0.8352 & \textbf{0.9197} \\
			\bottomrule[1pt]
		\end{tabular}%
				\end{table}
	\begin{table}
		\footnotesize
		\centering
		\caption{Training time (in second) on subsets of NUS-WIDE.}\setlength{\tabcolsep}{0.06cm}
			\begin{tabular}{p{1.8cm}<{\centering}|p{0.75cm}<{\centering}|p{0.88cm}<{\centering}|p{0.88cm}<{\centering}|p{0.9cm}<{\centering}|p{0.9cm}<{\centering}|p{1.4cm}<{\centering}}
			\toprule[1pt]
			Method & 1K    & 5K    & 10K   & 50K   & 184K  & Kernel time \\
			\hline
			CMFH~\cite{ding2016large}  & 43.03 & 345.14 & 977.17 & -     & -     & -  \\
			IMH~\cite{Song2013Inter}   & 0.59  & 21.33 & 140.26 & -     & -     & -  \\
			SePH\_km \cite{Lin2015Semantics} & 2.27  & 64.32 & 250.36 & -     & -     & 323.50 \\
			GSePH\_km \cite{Mandal2017Generalized} & 135.23 & 1343.94 & 4182.59 & -     & -     & 352.82 \\
			DCH~\cite{Xu2017Learning}   & 1.08  & 2.15  & 4.45  & 40.21 & 242.84 & -  \\
			DLFH~\cite{8636536}  & 21.20 & 86.92 & 186.34 & 1071.59 & 3701.65 & -  \\
			SCRATCH~\cite{8691805}  & 1.27  & 4.83  & 8.68  & 39.98 & 141.12 & 40.48 \\
SRLCH~\cite{shen2020exploiting}  & 3.55 &     3.86 &      5.01 &      14.21 &   44.43 &     40.05 \\
BATCH~\cite{BATCH}  &  3.16 &       3.08 &     3.41 &     7.47 &     20.78 &    40.82 \\
			\hline
			FDDH  & \textbf{0.25}  & \textbf{0.83}  & \textbf{1.74}  & \textbf{6.14}  & \textbf{19.75} & \textbf{40.01} \\
			\bottomrule[1pt]
		\end{tabular}%
		\label{tab:addlabel5}%
	\end{table}%

\begin{figure*}[!t]
\begin{center}
	\includegraphics[width=18.2cm]{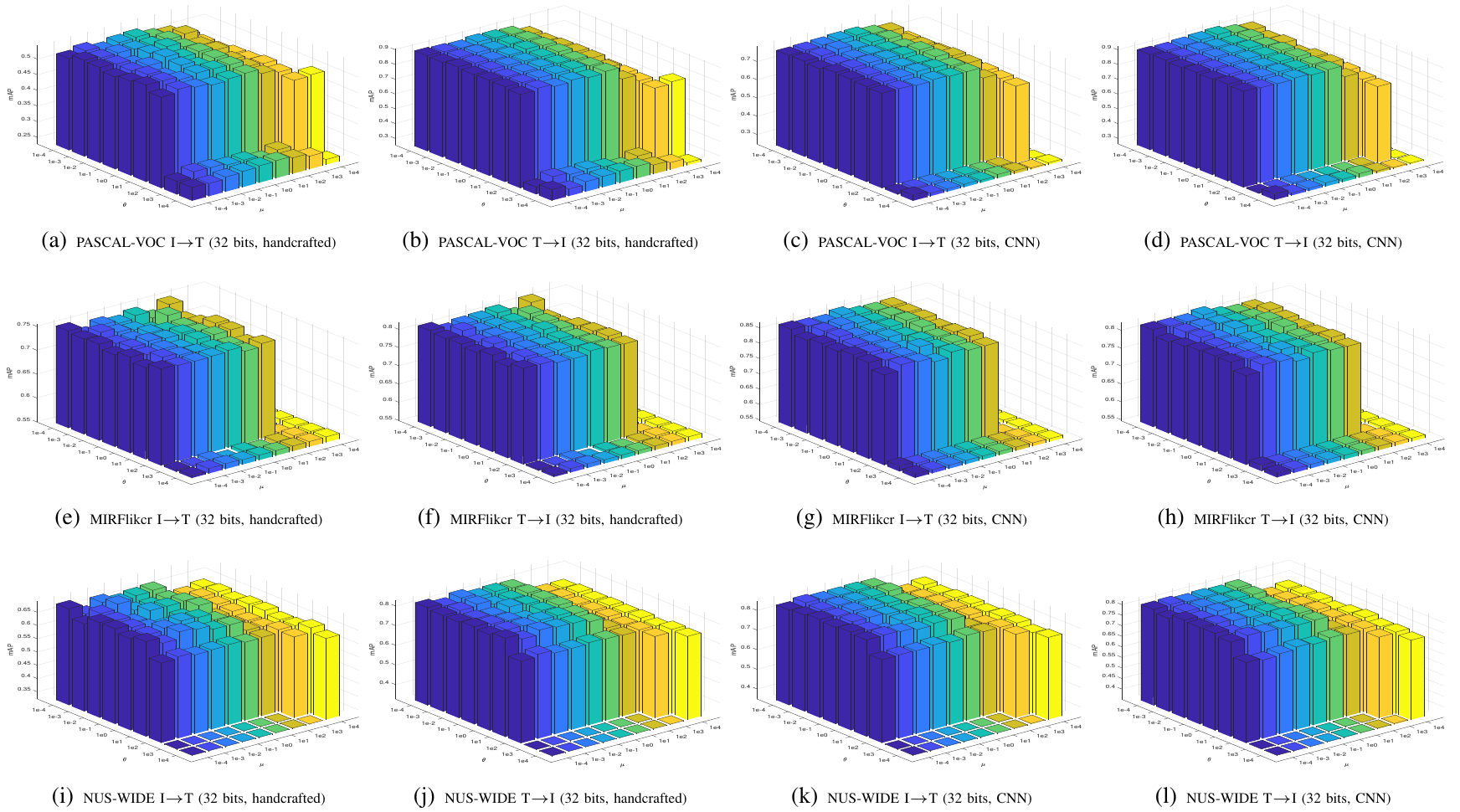}
\end{center}
\vspace{-0.2cm}	
\caption{The sensitivity analysis of the hyper-parameters $\mu$ and $\theta$  tested on handcrafted and CNN visual features.}
		\label{fig:figure5}
	\end{figure*}

	3)~\emph{\textbf{Results of Training Time}}: The computational complexity of the proposed FDDH
framework mainly accumulates from the training process, in which the processing time of RBF kernel mapping  is   a constant during the learning process.  For fair comparison, we evaluate the competing algorithms on NUS-WIDE dataset with handcrafted features, and  record the execution time of different training sizes with 128 hash bits.
It is noted that the representative CMFH, IMH, SePH and GSePH method are computationally intractable on large-scale training dataset, and we only implement these methods on relatively small training size, \emph{i.e.}, the data sizes are less than 10k. The training times tested on different data sizes and conducted on different methods are shown
in Table~\ref{tab:addlabel5}, it can be observed that the unsupervised CMFH and supervised GSePH methods often require large training time to learn the hash codes. The main reason lies that CMFH often involves large iterations to convergence, while GSePH approach takes massive amount of computations to factorize the large affinity matrix. Meanwhile,
the deep learning method, \emph{i.e.}, DLFH, often requires large training  time to perform the learning process. Evidently, these methods could be quite time-consuming when the training database is too
large.

Comparatively speaking, the competing IMH, DCH, SCRATCH, SRLCH, BATCH and the proposed FDDH methods have significantly reduced the training time.
Since the proposed FDDH approach considers more variables to discriminatively learn the hash codes, the execution time of training time could be much higher than the relevant
IMH, DCH, SCRATCH, SRLCH and BATCH methods. Fortunately,  the proposed FDDH method not only significantly reduces the training time, but also achieves the best cross-modal retrieval performance in most cases. For instance,
the execution time obtained by the proposed FDDH method is only around 1.74m, in contrast to 140.26m, 4.45m , 8.68m, 5.01m and 3.41m, respectively, evaluated on IMH, DCH, SCRATCH, SRLCH and BATCH methods.
The main advantages contributed to such fast training process are three-fold: 1) The utilization of  orthogonal constraint to  hash code learning can well reduce the quantization error, which therefore speed up the learning process. 2) The $\varepsilon$-dragging technique is able to provide large class margins for fast regression, which
 can make the optimization procedure converge within less iterations. 3) The proposed FDDH framework has a close-form solution to hash code learning and only requires a single step
to update the whole hash codes. As a result, the FDDH algorithm performs sufficiently fast to process the large-scale database.

\begin{figure}[!t]
\begin{center}
	\includegraphics[width=9cm]{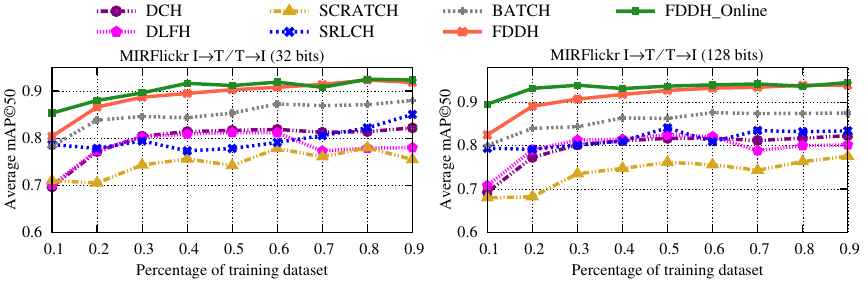}
\end{center}
\vspace{-0.3cm}	
\caption{Retrieval results tested with different training sizes.}
		\label{fig:figure6}
	\end{figure}

	4)~\emph{\textbf{Effective of Online Strategy}}: For the out-of-sample extensions, the offline and online strategy are both proposed to learn the modality-specific projections.
Similar to most competing works,  the offline learning strategy is selected as the standard way for learning. Besides,  an efficient online learning strategy is newly proposed to adapt the streaming data.
For simplicity, the proposed FDDH algorithm associated with online learning strategy is abbreviated as FDDH\_online. Further, we vary the training size of  MIRFlickr dataset
 from 10\% to 90\% and select the handcrated features of MIRFlickr dataset for illustration. Accordingly, the average mAP@50 scores tested on I${\rightarrow}$T and T${\rightarrow}$I tasks are
  shown in the Fig.~\ref{fig:figure6}, it can be observed that both of the proposed FDDH and FDDH\_online methods always outperform other state-of-art competing methods on
  different training sizes. Remarkably, the proposed FDDH\_online scheme has yielded very promising retrieval performance when the training size is relatively small (less than 30\%).
This impressive achievement can be attributed to the stable ability of the online learning scheme illustrated in
Section~\ref{online}, which can adaptively learn the modality-specific projections within the different training sizes.
Therefore, the proposed online learning scheme is able to well learn discriminative hash codes with limited training dataset,
which can be efficiently utilized to facilitate large-scale cross-modal retrieval tasks.

	5)~\emph{\textbf{Effects of Model Parameters}}: There are three main parameters involved within the proposed FDDH framework, \emph{i.e.},  $\mu, \theta, \delta$.
 Specifically, $\delta$ is the weight coefficient to control the degree of semantic label relaxation, and the extensive experiments find the retrieval performance are always stable
 when the value of $\delta$ is within the range of  ${[10^2, 10^3]}$. Meanwhile,  we fix the code length to be 32 bits and set $\delta$ at ${10^2}$ to
further attempt different values of  $\mu$ and $\theta$, by varying the value of one parameter while fixing the
another one.  More specifically, we vary the parameter $\mu$ within $\{10^{-4},...,10^{4}\}$ and $\theta$ within $\{10^{-4},...,10^{4}\}$.
  The mAP scores tested on different $\{\mu,\theta\}$ values and evaluated on different retrieval tasks are shown in Fig.~\ref{fig:figure5}, it can be observed that
the mAP scores obtained by the FDDH are very stable in most cases, and the  performance deteriorates only when the $\{\mu,\theta\}$  values are very large, \emph{e.g.} $\mu$ greater than $10$ and $\theta$ greater than $10^{3}$,
Overall, the results perform well when $\mu$ is selected within the range of $[10^{-4},10]$ and $\theta$ is chosen within the range of $[10^{-4},10^{3}]$. Besides, the different  $\{\mu,\theta\}$ values only induce
a minor fluctuation to the retrieval performance. Therefore,
these parameters are generally insensitive to the cross-modal
retrieval performances within a wide range of values.

\begin{figure}[!t]
\begin{center}
	\includegraphics[width=9cm]{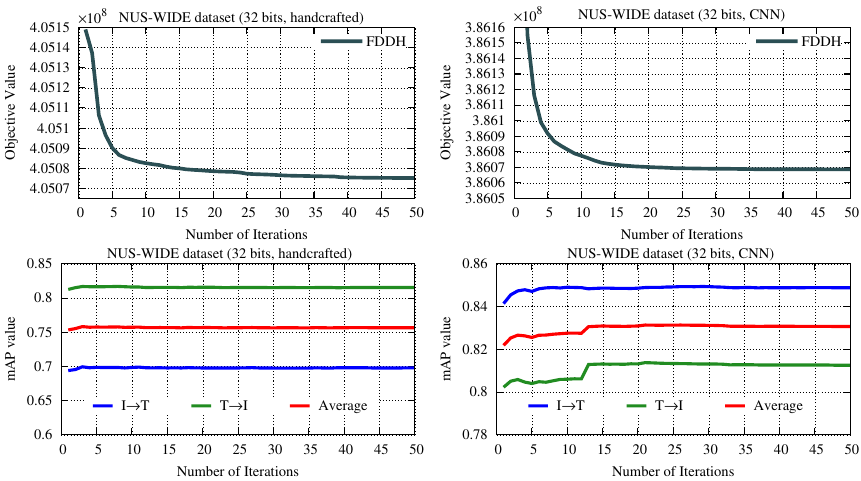}
\end{center}
	\vspace{-0.3cm}	
	\caption{Results of objective function values and mAP scores recorded in different iterations and tested on NUS-WIDE dataset.}
		\label{fig:figure7}
	\end{figure}

	6)~\emph{\textbf{Convergence Analysis}}: Algorithm \ref{alg:1} shows the main procedures of FDDH algorithm, and its
 complexity has been theoretically analyzed in Section~\ref{theory}. Besides, we further study the convergence of the proposed FDDH algorithm.
 By fixing the code length to be 32 bits, we record the objective value and mAP score at each iteration and select NUS-WIDE dataset for illustration.  As shown in Fig.~\ref{fig:figure7}, it can be observed that the FDDH converges very fast during the learning process. On the one hand, the objective values almost converge within 30 and 20 iterations, respectively, tested on handcrafted and CNN visual features. On the other hand, the corresponding mAP scores converge to a stable value within very limited iterations, \emph{e.g.}, 5 iterations for handcrafted features and 15 iterations for CNN visual features, respectively.
 Therefore, the overall computational load of the proposed FDDH can be significantly reduced due to the less iterations.

%
\begin{table}[!t]
		\centering
		\caption{\footnotesize{Different optimizations tested on NUS-WIDE dataset (128bits).}}
		\begin{tabular}{c|c|c|c|c|c}
			\toprule[1pt]
			\multirow{1}[2]{*}{Method} & \multicolumn{2}{c|}{MAP} & \multicolumn{2}{c|}{Top50 precison} & \multirow{1}[2]{*}{Training time} \\
			\cline{2-5}          & I${\rightarrow}$T   & T${\rightarrow}$I   & I${\rightarrow}$T   & T${\rightarrow}$I   & second (s) \\
			\hline
			DPLM-M & 0.4062 & 0.4300 & 0.5077 & 0.5701 & 126.23 (s) \\
			FSDH-M & 0.6511 & 0.7961 & 0.7712 & 0.8745 & 96.05 (s) \\
			\hline
			FDDH-DPLM & 0.4620 & 0.4791 & 0.6852 & 0.7364 & 102.63 (s) \\
			FDDH & \textbf{0.7118} & \textbf{0.8244} & \textbf{0.8596} & \textbf{0.9234} & \textbf{59.76} (s) \\
			\bottomrule[1pt]
		\end{tabular}%
		\label{tab:addlabel6}%
	\end{table}%

	7)~\emph{\textbf{Different Optimizations}}: In the literature, several fast unimodal hashing methods have been presented, \emph{e.g.}, DPLM~\cite{Shen2016A} and FSDH~\cite{Gui2018Fast}.
Similar to the multi-modal extension from SDH~\cite{Shen2015CVPR} to DCH~\cite{Xu2017Learning}, we heuristically extend DPLM and FSDH to adapt the multi-modal data (respectively abbreviated
as DLPM-M and FSDH-M), and compare the proposed FDDH approach with these fast hashing schemes.
Specifically, the large NUS-WIDE dataset associated with handcrafted features is selected for illustration, and  we also utilize the optimization
	scheme within DPLM to update $\mathbf{H}$ in Eq.~(\ref{eq6}) (abbreviated as FDDH-DPLM).
Accordingly, we record the retrieval performances and training times by different optimizations in Table~\ref{tab:addlabel6},
it can be found that the proposed FDDH approach not only produces the highest mAP scores and top-50 precisions, but also involves the lowest training time.
It is noted that the DPLM-M and FSDH-M methods respectively share the similar optimization algorithms with
FDDH-DPLM and FDDH to achieve cross-modal retrieval
tasks, and it is easy to find that the results of FDDH-DPLM
and FDDH involve the lower training times, while producing
the better retrieval performances. The main reason lies that the proposed FDDH method imposes an orthogonal constraint to reduce the quantization error, while employing the relaxed semantic label values for fast convergence.
 Remarkably, the  optimization algorithm within FDDH framework significantly speeds up the learning process and surprisingly contributes to the fastest implementation. Therefore, the proposed FDDH algorithm runs sufficiently fast, and it is particularly suitable for processing the large-scale multimedia datasets.

\section{Conclusion}\label{conclusion}
This paper has proposed a novel fast discriminative discrete hashing method for large-scale cross-modal retrieval. The
proposed framework introduces an orthogonal basis to regress the targeted hash codes  of training examples
to their corresponding reasonably relaxed class label values, which  offers provable large margin property to efficiently
reduce the quantization error. Meanwhile, an orthogonal transformation
scheme is further proposed to guarantee the semantic consistency between the data feature vector and its semantic
representation. Through the joint exploitation of the above, an efficient closed-form solution is derived for discriminative hash code learning, while an effective
online strategy is newly proposed for modality-specific projection function learning. Extensive experiments empirically show that the proposed FDDH
approach performs sufficiently fast, and  brings substantial improvements over the state-of-the-art methods.



\begin{IEEEbiography}[{\includegraphics[width=1in,height=1.25in,clip,keepaspectratio]{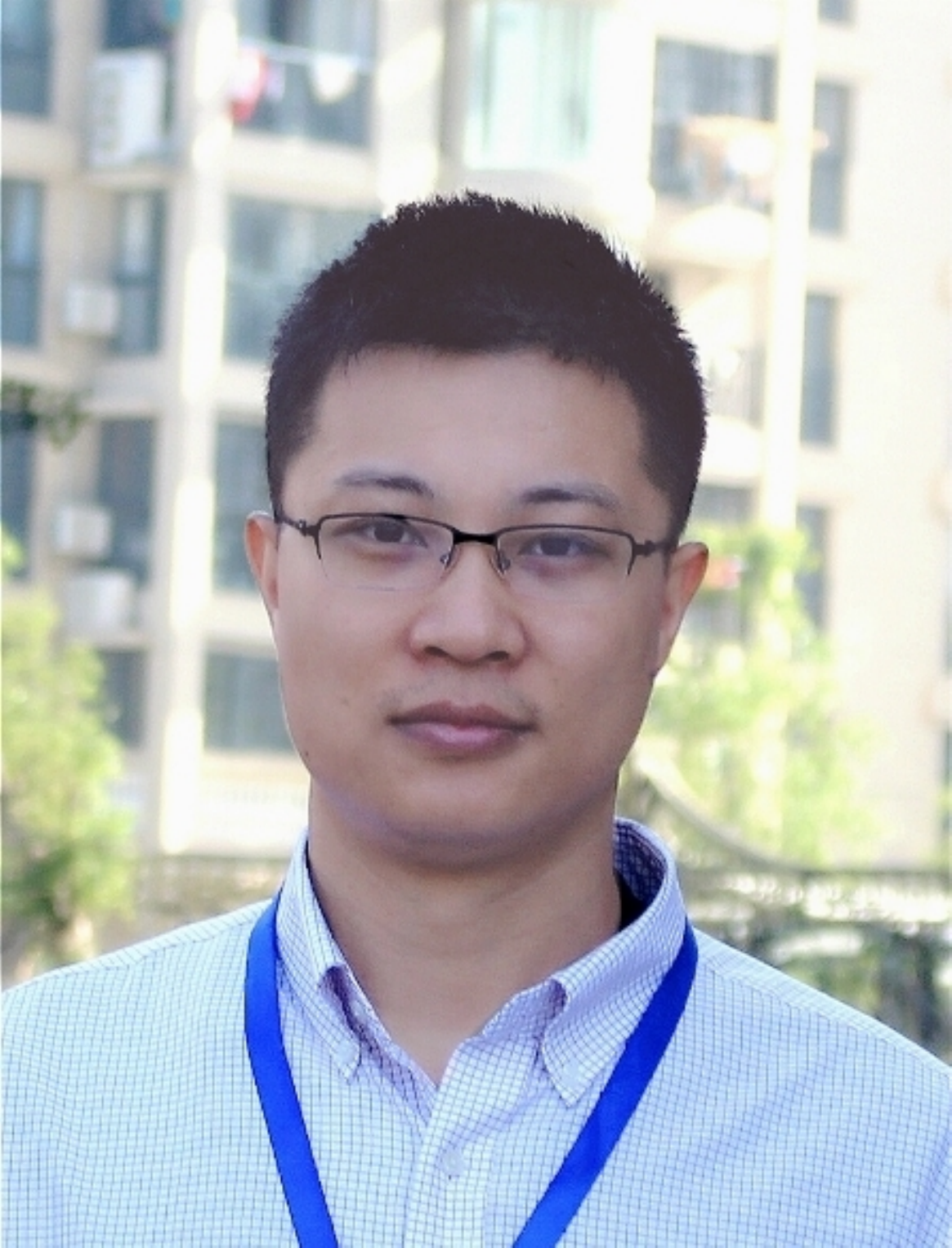}}]{Xin Liu}
received the Ph.D. degree in computer
science from Hong Kong Baptist University, Hong Kong, in 2013. He was a visiting scholar with  Computer \& Information Sciences Department, Temple University, Philadelphia, USA, from 2017 to 2018. Currently, he is
a Full Professor with the Department of Computer Science and Technology, Huaqiao University, Xiamen, China, and also a Research Fellow with the
State Key Laboratory of Integrated Services Networks of Xidian University, China.  His present research interests include multimedia analysis, computational intelligence, computer vision, pattern recognition and machine learning.  He is a senior member of the IEEE.
\end{IEEEbiography}
\begin{IEEEbiography}[{\includegraphics[width=1in,height=1.25in,clip,keepaspectratio]{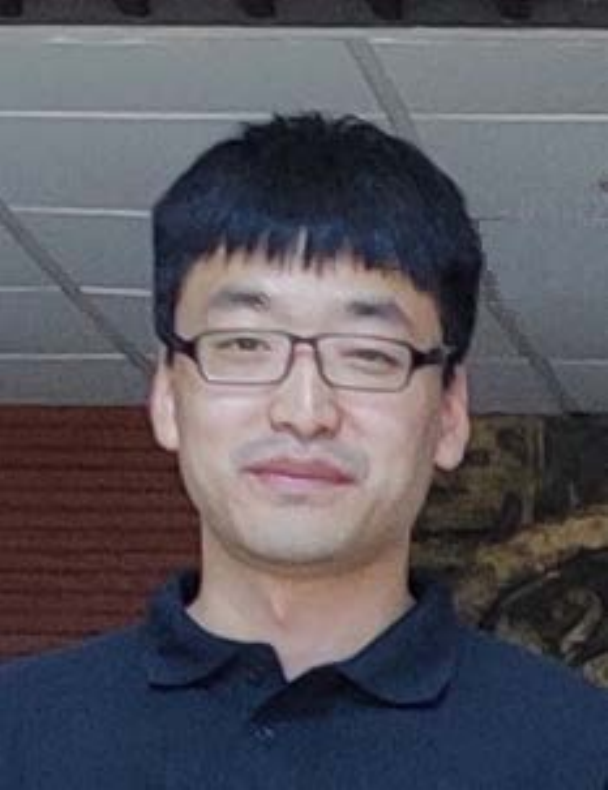}}]{Xingzhi Wang} received the  MS degree in computer
science from Huaqiao University, Xiamen, China,
in 2020. He is currently a PHD student at School of Electrics and Information Technology, Sun Yat-sen University, Guangzhou, China. His present research interests include multimedia signal processing, pattern recognition and affective computing. He is a student member of IEEE.
\end{IEEEbiography}
\begin{IEEEbiography}[{\includegraphics[width=1in,height=1.25in,clip,keepaspectratio]{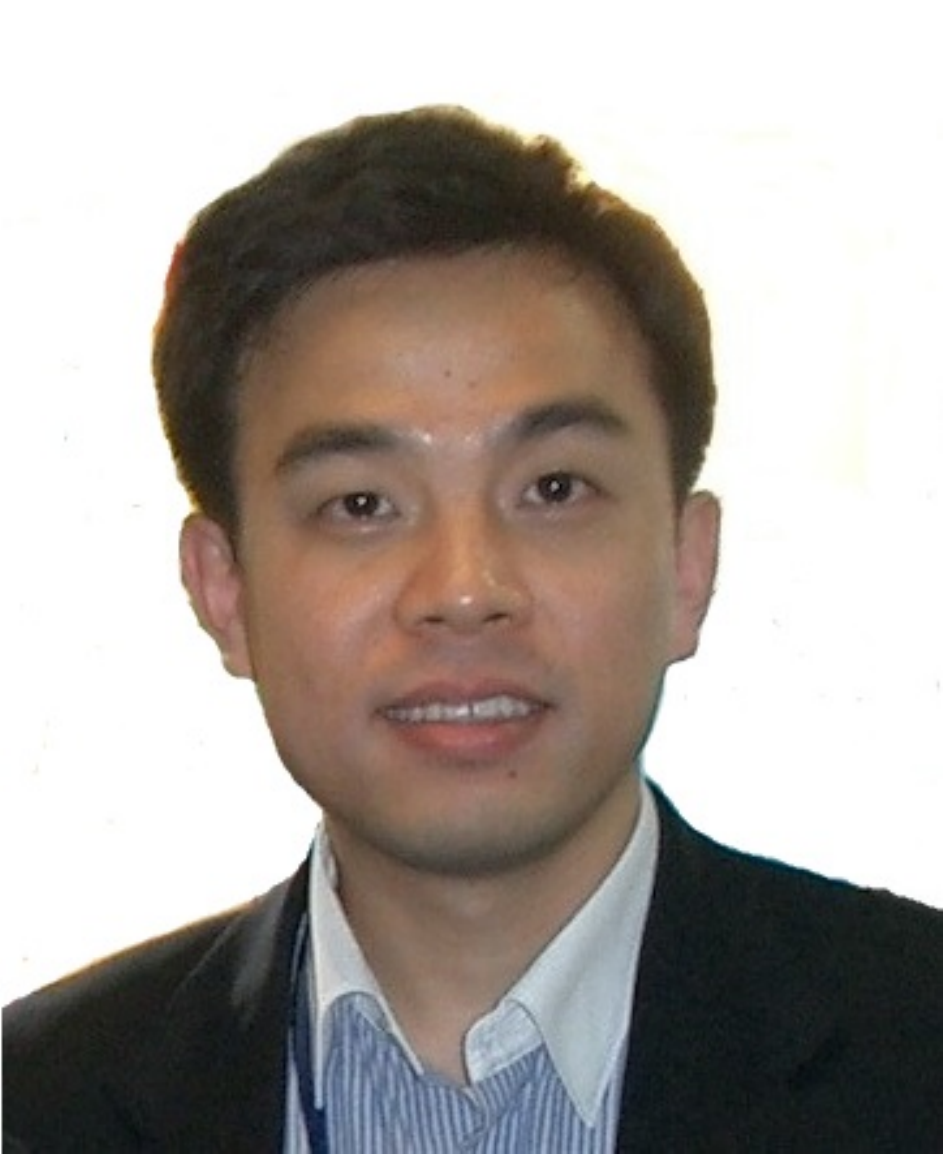}}]{Yiu-ming Cheung}
received his Ph.D. degree from the Department of Computer
Science and Engineering, Chinese University of
Hong Kong, Hong Kong. He is currently a Full Professor with the
Department of Computer Science, Hong Kong
Baptist University, Hong Kong. His current
research interests include machine learning, pattern
recognition and visual computing.
Prof. Cheung is the Founding Chairman of the Computational Intelligence Chapter of the
IEEE Hong Kong Section. He serves as an Associate Editor for the IEEE Transactions on Neural Networks and Learning Systems, IEEE Transactions on Cybernetics, Pattern Recognition,
 Knowledge and Information Systems, Neurocomputing, to name a few. He is an IEEE Fellow, IET Fellow, RSA Fellow and BCS Fellow.
\end{IEEEbiography}

	\clearpage
	\begin{appendix}
In this part,  we provide a detailed proof for stability concerning to the online learning strategy. For the new stream data $\mathbf{X}_s^t$ and $\mathbf{X}_s^{t/i}$ (only the $i$-th vector is different), the following  relationship is stabilized within the online strategy:
	\begin{equation}	
 {\left\| {{\mathbf{P}^t}{(\mathbf{X}_s^t)} - {\mathbf{P}^t}{({\mathbf{X}}_s^{t/i}})} \right\|_F} \le {\mathbf{\Theta}(n)} \nonumber
	\label{eq30}
	\end{equation}
	where ${\mathbf{\Theta}(n)}$ converges to zero as $n$ goes to infinity.
	Before the proof,  Bregman matrix divergence is introduced.



\textbf{\textit{Bregman Divergence}}: For any matrix $\mathbf{A}$ and $\mathbf{B}$ of same size, the expression of Bregman divergence with respect to function $f$ is defined as follows:
    	\begin{equation} 		
    	\texttt{Bgm}_f(\mathbf{A},\mathbf{B}) = f(\mathbf{A}){-}f(\mathbf{B}){-}\text{Tr}\left( {\nabla f{{(\mathbf{B})}^{\rm{T}}}(\mathbf{A}{-}\mathbf{B})} \right) \nonumber
		\label{eq33}    	
		\end{equation} 		
		where ${\nabla f(\mathbf{B})}$ represents the derivative of $f$ at matrix $\mathbf{B}$. The two properties of Bregman Divergence are as follows:
		
		\textit{1) Non-Negative property}: If the function $f$ is convex,
    	 Bregman divergence  conforms to the non-negative  property:
    	\begin{equation} 			
    		\texttt{Bgm}_f(\mathbf{A},\mathbf{B})  \ge 0 \nonumber
		\end{equation} 	

        	\textit{2) Additive Property}: If the functions $f$ and $g$ are convex, Bregman divergence also satisfies the additive property:
        	\begin{equation} 			
			\texttt{Bgm}_{f{+}g}(\mathbf{A},\mathbf{B}) = \texttt{Bgm}_f(\mathbf{A},\mathbf{B}) + \texttt{Bgm}_g(\mathbf{A},\mathbf{B}) \nonumber
		\end{equation} 	

		
Before presenting our proof,  the objective function is modified to ensure that the regularization parameters
are invariant to the existing sample size $n$, new data size $m$, the dimensionality $d$ of RBF mapping  and hash code length $q$:
		\begin{equation}
		\begin{aligned}
{G_{\tiny{\rm{online}}}}({{\mathbf{P}}_t}) &{=} \frac{1}{{q(n{+}m)}}\left\| {{\mathbf{H}}{-}{{\mathbf{P}}_t}\phi ({{\mathbf{X}}^{t}})} \right\|_F^2{+}\frac{\gamma }{{qd}}\left\| {{{\mathbf{P}}_t}} \right\|_F^2 \\[1mm]
		& +\frac{1}{{q(n{+}m)}}\left\| {\mathbf{H}_s{-}{{\mathbf{P}}_t}\phi ({\mathbf{X}}_s^{t})} \right\|_F^2
		\end{aligned}
		\label{eq31}
		\end{equation}

		
Without loss of generality, we assume that the optimization stops
within $K$ iterations and in the $k$-th iteration ($k{=}1,\cdots,K)$, $\mathbf{H}^k$, $\mathbf{H}_s^k$ and $\mathbf{P}_t^k$ are obtained. In the following, we  prove that the learning of $\mathbf{P}_t$ is stable in each iteration.

\textbf{\textit{Theorem 1}}: Let  $\mathbf{X}^t$, $\mathbf{H}$  and $\mathbf{X}_s^t$, respectively, denote the training dataset, hash code matrix and new data samples of the $t$-th modality,
${\mathbf{P}_t}(\mathbf{X}^t_s)$ and ${\mathbf{P}_t}(\mathbf{X}_s^{t/i})$ represent the modality-specific projection results acted directly on the samples $\mathbf{X}_s^t$ and $\mathbf{X}_s^{t/i}$
in the $k$-th iteration, the online strategy defined in Eq.~(\ref{eq31}) is stable when learning ${\mathbf{P}_t}$. In addition, for any already learned ${\mathbf{H}}$ and ${\mathbf{P}_t}$, we assume
that ${\left\| {{\mathbf{h}}{-}{\mathbf{P}_t}\phi ({\mathbf{x}})} \right\|_2}{\le}{\rm{M}}$, where ${\rm{M}}$ is a constant. Then,
		\begin{equation} 		
		{\left\| {{{\mathbf{P}_t}^k}({\mathbf{X}^t_s}) - {{\mathbf{P}_t}^k}({\mathbf{X}_s^{t/i}})} \right\|_F} \le \frac{{2{d^2}{\rm{M}}}}{{\gamma (n{+}m)}}
		\label{eq32}
		\end{equation} 		
		\textbf{\textit{Proof of the Theorem 1}}: In the $k$-th iteration, let
		\begin{equation}
		\begin{aligned}			
{f_{\mathbf{X}}}({\mathbf{P}_t})&{=}\frac{1}{{q(n{+}m)}}(\left\| {{\mathbf{H}}{-}{\mathbf{P}_t}\phi ({\mathbf{X}^t})} \right\|_F^2 + \left\| {{\mathbf{H}_s^t}{-}{\mathbf{P}_t}\phi ({{\mathbf{X}}_s^t})} \right\|_F^2)\\[1mm]\nonumber
&+ \frac{\gamma }{{qd}}\left\| {\mathbf{P}_t} \right\|_F^2
		\end{aligned}
		\end{equation} 	
		and
		\begin{equation}
		{{{g}}_{\mathbf{X}}}({\mathbf{P}_t}){\rm{ = }}\frac{\gamma }{{qd}}\left\| {\mathbf{P}_t} \right\|_F^2 \nonumber
		\end{equation} 			
		Considering to the non-negative and additive properties of Bregman divergence, we can obtain the following property:
		\begin{equation}
		\begin{array}{l}
		{\texttt{Bgm}_{f_{{{\mathbf{X}_s^{t/i}}}}}}\left( {{{\mathbf{P}}^k}({\mathbf{X}_s^{t}}),{{\mathbf{P}}^k}({{\mathbf{X}_s^{t/i}}})} \right){+}{\texttt{Bgm}_{f_{\mathbf{X}_s^{t}}}}\left( {{{\mathbf{P}}^k}({{\mathbf{X}}^i}),{{\mathbf{P}}^k}({\mathbf{X}})} \right)\\[1mm]\nonumber
		{\ge} {\texttt{Bgm}_{{g_{{{\mathbf{X}_s^{t/i}}}}}}}\left( {{{\mathbf{P}}^k}({\mathbf{X}_s^{t}}),{{\mathbf{P}}^k}({{\mathbf{X}_s^{t/i}}})} \right){+}{\texttt{Bgm}_{{g_{\mathbf{X}_s^{t}}}}}\left( {{{\mathbf{P}}^k}({{\mathbf{X}_s^{t/i}}}),{{\mathbf{P}}^k}({\mathbf{X}_s^{t}})} \right) 		\label{eq34}    	
		\end{array}
		\end{equation}
One the one hand, the extensions of  the above right parts  are:
		\begin{equation}
		\begin{array}{l}
		{\texttt{Bgm}_{g_{{{\mathbf{X}_s^{t/i}}}}}}\left( {{{\mathbf{P}}^k}({\mathbf{X}_s^{t}}),{{\mathbf{P}}^k}({\mathbf{X}_s^{t/i}})} \right){+} {\texttt{Bgm}_{{g_{\mathbf{X}_s^{t}}}}}\left( {{{\mathbf{P}}^k}({\mathbf{X}_s^{t/i}}),{{\mathbf{P}}^k}({\mathbf{X}_s^{t}})} \right)\\[1mm]\nonumber
		= \frac{\gamma }{{qd}}\left\| {{{\mathbf{P}}^k} ({\mathbf{X}_s^{t}})} \right\|_F^2 - \frac{\gamma }{{dq}}\left\| {{{\mathbf{P}}^k}({{\mathbf{X}}_s^{t/i}})} \right\|_F^2\\[1mm]\nonumber
		- \frac{{2\gamma }}{{qd}}tr\left( {{{\mathbf{P}}^k}{{({{\mathbf{X}}_s^{t/i}})}^{\rm{T}}}\left( {{{\mathbf{P}}^k} ({\mathbf{X}_s^{t}}) - {{\mathbf{P}}^k}({{\mathbf{X}}_s^{t/i}})} \right)} \right)\\[1mm]\nonumber
		{+}\frac{\gamma }{{qd}}\left\| {{\mathbf{P}^k}({\mathbf{X}_s^{t/i}})} \right\|_F^2 - \frac{\gamma }{{dq}}\left\| {{{\mathbf{P}}^k}({\mathbf{X}_s^{t}})} \right\|_F^2\\[1mm]\nonumber
		- \frac{{2\gamma }}{{qd}}tr\left( {{{\mathbf{P}}^k}{{({\mathbf{X}_s^{t}})}^{\rm{T}}}\left( {{{\mathbf{P}}^k}({\mathbf{X}_s^{t/i}})- {{\mathbf{P}}^k}({\mathbf{X}_s^t})} \right)} \right)\\[1mm]\nonumber
		= \frac{{2\gamma }}{{qd}}\left\| {{{\mathbf{P}}^k}({\mathbf{X}_s^{t}})- {{\mathbf{P}}^k}({{\mathbf{X}}_s^{t/i}})} \right\|_F^2
		\label{eq35}    	
		\end{array}
		\end{equation} 	
		
On the other hand,  the optimal solution of $\mathbf{P}_t$ is obtained by setting the derivative of Eq.~\eqref{eq19} to zero. Therefore, we can obtain  $\nabla f{{(\mathbf{P}_t^k)}}=0$. Since  $\mathbf{x}$ is  zero-centered sample, it can be concluded that $\left\| \phi(\mathbf{x})\right\|_2\leq{d}$  and we have:
		\begin{equation}
		\begin{array}{l}
		{\texttt{Bgm}_{f_{{{\mathbf{X}_s^{t/i}}}}}}\left( {{{\mathbf{P}}^k}({\mathbf{X}_s^t}),{{\mathbf{P}}^k}({{\mathbf{X}_s^{t/i}}})} \right) + {\texttt{Bgm}_{f_{\mathbf{X}_s^t}}}\left( {{{\rm{P}}^k}({{\mathbf{X}}_s^{t/i}}),{{\mathbf{P}}^k}({\mathbf{X}_s^t})} \right)\\[2mm]\nonumber
		= \frac{1}{{q(n{+}m)}}\left( {\left\| {{\mathbf{H}}{-}{{\mathbf{P}}^k}({{\mathbf{X}}_s^t})\phi ({\mathbf{X}^t})} \right\|_F^2{+}\left\| {{\mathbf{H}}_s^k - {{\mathbf{P}}^k}({{\mathbf{X}}_s^t})\phi ({{\mathbf{X}}_s^{t/i}})} \right\|_F^2} \right)\\[2mm]\nonumber
		+ \frac{\gamma }{{qd}}\left\| {{{\mathbf{P}}^k}({{\mathbf{X}}_s^t})} \right\|_F^2 - \frac{1}{{q(n{+}m)}}\left\| {{\mathbf{H}} - {{\mathbf{P}}^k}({\mathbf{X}}_s^{t/i})\phi ({\mathbf{X}^t})} \right\|_F^2\\[2mm]\nonumber
		- \frac{1}{{q(n{+}m)}}\left\| {{\mathbf{H}}_s^k - {{\mathbf{P}}^k}({\mathbf{X}}_s^{t/i})\phi ({\mathbf{X}_s^{t/i}})} \right\|_F^2 - \frac{\gamma }{{dq}}\left\| {{{\mathbf{P}}^k}({\mathbf{X}}_s^{t/i})} \right\|_F^2\\[2mm]\nonumber
		{+}\frac{1}{{q(n{+}m)}}\left( {\left\| {{\mathbf{H}}{-}{{\mathbf{P}}^k}({\mathbf{X}}_s^{t/i})\phi ({\mathbf{X}^t})} \right\|_F^2 + \left\| {{\mathbf{H}}_s^k - {{\mathbf{P}}^k}({\mathbf{X}}_s^{t/i})\phi ({\mathbf{X}_s^t})} \right\|_F^2} \right)\\[2mm]\nonumber
		+ \frac{\gamma }{{qd}}\left\| {{{\mathbf{P}}^k}({\mathbf{X}}_s^{t/i})} \right\|_F^2 - \frac{1}{{q(n{+}m)}}\left\| {{\mathbf{H}} - {{\mathbf{P}}^k}({{\mathbf{X}}_s^t})\phi ({\mathbf{X}^t})} \right\|_F^2\\[2mm]\nonumber
		- \frac{1}{{q(n{+}m)}}\left\| {{\mathbf{H}}_s^k - {{\mathbf{P}}^k}({{\mathbf{X}}_s^t})\phi ({\mathbf{X}_s^t})} \right\|_F^2 - \frac{\gamma }{{dq}}\left\| {{{\mathbf{P}}^k}({{\mathbf{X}}_s^t})} \right\|_F^2\\[2mm]\nonumber
		{=}\frac{1}{{q(n{+}m)}}\left( {\left\| {{\mathbf{H}}_s^k{-}{{\mathbf{P}}^k}({{\mathbf{X}}_s^t})\phi ({\mathbf{X}}_s^{t/i})} \right\|_F^2{-}\left\| {{\mathbf{H}}_s^k{-} {{\mathbf{P}}^k}({\mathbf{X}}_s^{t/i})\phi ({\mathbf{X}}_s^{t/i})} \right\|_F^2} \right)\\[2mm]\nonumber
		{+}\frac{1}{{q(n{+}m)}}\left( {\left\| {{\mathbf{H}}_s^k - {{\mathbf{P}}^k}({\mathbf{X}}_s^{t/i})\phi ({{\mathbf{X}}_s^t})} \right\|_F^2{-}\left\| {{\mathbf{H}}_s^k - {{\mathbf{P}}^k}({{\mathbf{X}}_s^t})\phi ({{\mathbf{X}}_s^t})} \right\|_F^2} \right)\\[2mm]\nonumber
		= \frac{1}{{q(n{+}m)}}\left( {\left\| {{\mathbf{h}}_i^k{-}{{\mathbf{P}}^k}({{\mathbf{X}}_s^t})\phi ({\mathbf{x}_{i'}^{s}})} \right\|_2^2{-}\left\| {{\mathbf{h}}_i^k - {{\mathbf{P}}^k}({\mathbf{X}}_s^{t/i})\phi ({\mathbf{x}}_{i'}^{s})} \right\|_2^2} \right)\\[2mm]\nonumber
		+ \frac{1}{{q(n{+}m)}}\left( {\left\| {{\mathbf{h}}_i^k - {{\mathbf{P}}^k}({\mathbf{X}}_s^{t/i})\phi ({{\mathbf{x}}_i^s})} \right\|_2^2{-}\left\| {{\mathbf{h}}_i^k - {{\mathbf{P}}^k}({\mathbf{X}}_s^{t})\phi ({{\mathbf{x}}_i^s})} \right\|_2^2} \right)\\[2mm]\nonumber
		\le \frac{{2{\rm{M}}}}{{q(n{+}m)}}\left( {{{\left\| {{\mathbf{h}}_i^k - {{\mathbf{P}}^k}({{\mathbf{X}}_s^t})\phi ({\mathbf{x}}_{i'}^{s})} \right\|}_2} - {{\left\| {{\mathbf{h}}_i^k - {{\mathbf{P}}^k}({\mathbf{X}}_s^{t/i})\phi ({\mathbf{x}}_{i'}^{s})} \right\|}_2}} \right)\\[2mm]\nonumber
		+ \frac{{2{\rm{M}}}}{{q(n{+}m)}}\left( {{{\left\| {{\mathbf{h}}_i^k - 	{{\mathbf{P}}^k}({{\mathbf{X}}_s^{t/i}})\phi ({{\mathbf{x}}_i^s})} \right\|}_2} - {{\left\| {{\mathbf{h}}_i^k - {{\mathbf{P}}^k}({\mathbf{X}}_s^t)\phi ({{\mathbf{x}}_s^i})} \right\|}_2}} \right)\\[2mm]\nonumber
		{\le}\frac{{2{\rm{M}}}}{{q(n{+}m)}} {{{\left\| {\left( {{{\mathbf{P}}^k}({{\mathbf{X}}_s^t}){-}{{\mathbf{P}}^k}({\mathbf{X}}_s^{t/i})}\right)\phi({\mathbf{x}_{i'}^{s}})} \right\|}_2}}
		\\[2mm]\nonumber
{+}\frac{{2{\rm{M}}}}{{q(n{+}m)}}{\left\| {\left( {{{\mathbf{P}}^k}({{\mathbf{X}}_s^t}){-}{{\mathbf{P}}^k}({\mathbf{X}}_s^{t/i})} \right)\phi ({\mathbf{x}}_{i}^{s})} \right\|_2}\\[2mm]\nonumber
		\le \frac{{4d{\rm{M}}}}{{q(n{+}m)}}{\left\| {{{\mathbf{P}}^k}({{\mathbf{X}}_s^t}) - {{\mathbf{P}}^k}({\mathbf{X}}_s^{t/i})} \right\|_F}
		\end{array}
		\label{eq36}    	
		\end{equation} 	
Then, we have
		\begin{equation}
	\begin{array}{l}
	 	\frac{{2\gamma }}{{dq}}\left\| {{{\mathbf{P}}^k}({{\mathbf{X}}_s^{t}}){-}{{\mathbf{P}}^k}({\mathbf{X}}_s^{t/i})} \right\|_F^2 \le \frac{{4d{\rm{M}}}}{{q(n{+}m)}}{\left\| {{{\mathbf{P}}^k}({{\mathbf{X}}_s^t}) - {{\mathbf{P}}^k}({\mathbf{X}}_s^{t/i})} \right\|_F}\\[2mm]\nonumber
		~~~~~~~~~~~~~{\left\| {{{\mathbf{P}}^k}({{\mathbf{X}}_s^t}) - {{\mathbf{P}}^k}({\mathbf{X}}_s^{t/i})} \right\|_F} \le \frac{{2{d^2}{\rm{M}}}}{{\gamma(n{+}m)}}	
		\nonumber
	\end{array}
		\end{equation}
		The proof is complete.
			
	\end{appendix}

\end{document}